\documentclass{article}
\usepackage{eapo_iclr_style,times}

\usepackage{amsmath,amsfonts,bm}

\def\eqref#1{equation~\ref{#1}}

\def\1{\bm{1}}

\DeclareMathAlphabet{\mathsfit}{\encodingdefault}{\sfdefault}{m}{sl}
\SetMathAlphabet{\mathsfit}{bold}{\encodingdefault}{\sfdefault}{bx}{n}

\usepackage{amsmath,amssymb,amsthm}
\usepackage{booktabs}
\usepackage{array,tabularx}
\usepackage[font=small,labelfont=bf]{caption}
\usepackage{graphicx}
\usepackage{microtype}
\usepackage{multirow}
\usepackage{xcolor}
\usepackage{hyperref}
\hypersetup{hidelinks}
\usepackage{url}

\newtheorem{proposition}{Proposition}

\DeclareRobustCommand{\method}{\textsc{PyroAdapt}}
\newcommand{\dporanking}{binary-label RDPO}
\newcommand{\papertablefont}{\fontsize{9}{11}\selectfont}
\newcommand{\local}{\mathcal{D}_{\mathrm{local}}}
\newcommand{\extreme}{\mathcal{D}_{+}}
\newcommand{\dtrain}{\mathcal{D}_{\mathrm{train}}}
\newcommand{\xtarget}{\mathcal{X}_{\mathrm{target}}}

\title{\method{}: Adapting Wildfire Prediction under Spatial Heterogeneity and Temporal Shift}

\author{%
Enyi Jiang\textsuperscript{1,2} \quad Wu Sun\textsuperscript{3}\\[3pt]
\normalfont\small\textsuperscript{1}Siebel School of Computing and Data Science,\\
\normalfont\small University of Illinois at Urbana-Champaign\\
\normalfont\small\textsuperscript{2}Computer Science, Stanford University\\
\normalfont\small\textsuperscript{3}Odum School of Ecology, University of Georgia\\[3pt]
\normalfont\small\texttt{enyij2@illinois.edu, enyij@stanford.edu}\\
\normalfont\small\texttt{wusun@uga.edu}}

\iclrpreprintcopy

\begin{document}

\maketitle


\begingroup
\setlength{\parskip}{4pt plus 0.5pt minus 0.5pt}
\setlength{\textfloatsep}{9.5pt plus 2pt minus 2pt}
\setlength{\floatsep}{8pt plus 2pt minus 2pt}
\setlength{\intextsep}{9.5pt plus 2pt minus 2pt}
\setlength{\abovedisplayskip}{5.75pt plus 1.5pt minus 2.5pt}
\setlength{\belowdisplayskip}{5.75pt plus 1.5pt minus 2.5pt}
\setlength{\belowdisplayshortskip}{3.5pt plus 1.5pt minus 1.5pt}
\captionsetup{skip=6.5pt}
\makeatletter
\renewcommand{\section}{\@startsection{section}{1}{\z@}
  {-1.7ex plus -0.4ex minus -0.2ex}{1.2ex plus 0.25ex}
  {\large\sc\raggedright}}
\renewcommand{\subsection}{\@startsection{subsection}{2}{\z@}
  {-1.5ex plus -0.4ex minus -0.2ex}{0.7ex plus 0.2ex}
  {\normalsize\sc\raggedright}}
\renewcommand{\subsubsection}{\@startsection{subsubsection}{3}{\z@}
  {-1.2ex plus -0.4ex minus -0.2ex}{0.45ex plus 0.15ex}
  {\normalsize\sc\raggedright}}
\renewcommand{\paragraph}{\@startsection{paragraph}{4}{\z@}
  {1.05ex plus 0.35ex minus 0.2ex}{-1em}{\normalsize\bf}}
\makeatother

\begin{abstract}
Prediction of wildfire occurrence is a rare-event problem compounded by spatial heterogeneity and temporal distribution shift, as fire occurrences are vastly outnumbered by non-occurrences, and predictor--fire relationship varies across space and time.
Models trained on historical fire data may perform poorly under new conditions and require adaptation to the target distribution before operational use.
We propose \method{}, a pretrain--retrieve--rank framework that adapts a pretrained model to target conditions by retrieving historical locations with similar conditions and fine-tuning on the retrievals through risk ranking.
For spatial adaptation, we condition risk on terrain, ecoregion embeddings, and fire rates, accounting for spatial context in the retrieval, and learn risk ordering from same-day fire--nonfire cell pairs.
We compare direct ranking, residual pairwise DPO (RDPO), and selective ranking through a unified score-gap formulation that characterizes their gradient allocation. 
Over California (discretized into 666 0.25$^\circ$$\times$0.25$^\circ$ grid cells), these objectives raise daily average precision from 21.62\% for continued focal fine-tuning to 24.35--24.57\%, and Top5\% recall from 18.70\% to 22.11--22.79\%.
Under a fixed daily detection budget of 34 cells (5\% area), selective ranking captures 344 additional positive cell--days.
For fires in the top 5\%/10\%/20\% of dry matter consumption, selective ranking raises recall by 39.70/28.18/20.50 percentage points, respectively. 
Furthermore, rolling evaluations over Yosemite show that the gains from ranking persist under temporal distribution shift.
Together, these results show that \method{} prioritizes the most fire-prone locations under a daily budget constraint and detects more extreme fire events.
\end{abstract}

\section{Introduction}
\label{sec:introduction}

Climate change is fueling more extreme wildfires \citep{abatzoglou_2025}, with devastating impacts on health \citep{qiu_2025}, ecosystems \citep{byrne_2024}, and the economy \citep{wang_2020}.
Extreme wildfires are challenging to predict \citep{wang_2022}, as they emerge from the complex interplay of fire weather \citep{coen_2018, duane_2021, swain_2021}, terrain \citep{sharples_2012}, vegetation fuels \citep{rao_2023, swain_2021}, and human factors such as ignition and suppression \citep{keeley_2025, kreider_2024, miller_2020, wu_2023}. 
Climate change further shifts the spatial pattern, seasonality, and distribution of fire weather \citep{jones_2022} and fuel conditions \citep{baltzer_2021, ellis_2022, halofsky_2020}, so the predictor–fire relationships a model learns from historical data may not hold in the year it is deployed.


Predicting wildfire occurrence can be posed as scoring each location on each day for its propensity for fire, conditioned on environmental covariates and spatial context \citep{jain_2020,kondylatos_2022}.
Three properties of the data render this problem difficult.
First, positive labels (fire occurrences) are rare, as most location--days record no fire, yielding severe class imbalance and few positive labels relative to the volume of data \citep{he2009learning, chawla2002smote, liu2009learning}.
Second, the distribution of weather and fuel conditions changes from year to year \citep{abatzoglou2016impact, jones_2022}, so inputs at deployment may come from a shifted distribution.
Third, the mapping from covariates to risk varies across space, as terrain, vegetation, and ecoregion mediate how a given set of conditions translates into fire risk \citep{parisien_2009, hawbaker_2013, syphard_2024}.
These factors compound: a model must learn predictor--fire relationships across heterogeneous landscapes from relatively few positive labels while generalizing to conditions not represented in the training record.
Consequently, risk scores learned from past data may be poorly matched to the deployment year.

We introduce \method{}, a unified \emph{Pretrain--Retrieve--Rank} framework for batch-transductive adaptation using complete target-period covariates without target labels. Figure~\ref{fig:framework} summarizes the data flow. Historical focal pretraining initializes a risk model. Unlabeled target inputs retrieve $k$ nearest labeled historical examples in standardized weather and spatial covariate space \citep{cover1967nearest,khandelwal2020generalization}. Their labels define positive--negative pairs for fine-tuning without focal supervision, using a ranking loss that penalizes scoring positive examples below negative examples \citep{burges2005ranknet,joachims2002optimizing}.

\begin{figure}[t]
\centering
\includegraphics[width=\linewidth]{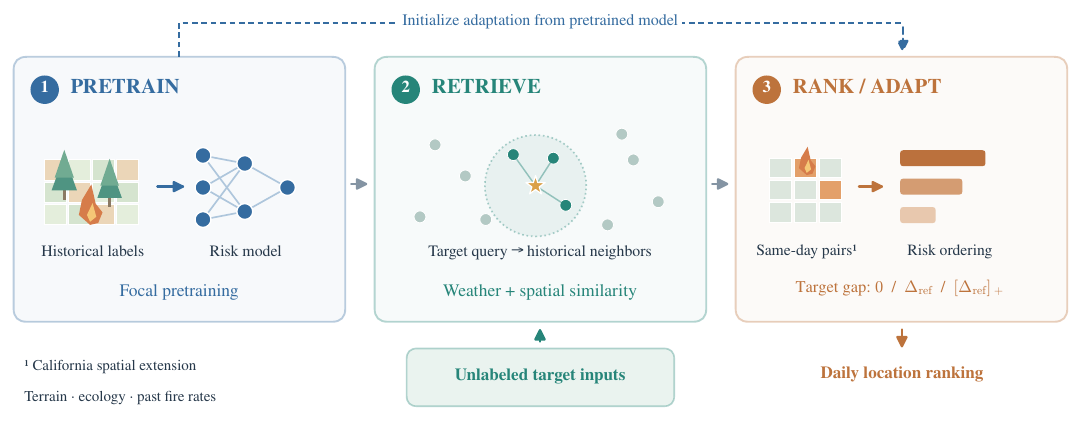}
\caption{\method{} overview. Given unlabeled target covariates, we retrieve their labeled historical neighbors (Step 2) and use the corresponding labels to adapt the pretrained model (Step 3). Target gaps (Step 3) correspond to direct ranking, RDPO, and selective ranking, respectively. For the California case, we add spatial covariates and same-day pairs for daily location ranking. Dashed arrows denote parameter initialization.}
\label{fig:framework}
\end{figure}


We analyze direct ranking, residual pairwise DPO (RDPO) \citep{rafailov2023dpo}, and selective ranking through a unified score-gap formulation. Direct ranking uses a zero target for adapted positive--negative score gaps; RDPO uses frozen reference gaps, and selective ranking retains pretrained targets only on correctly ordered pairs. Differentiating each loss with respect to the adapted gap reveals how these targets allocate gradients across correctly ordered and misordered pairs. For the \dporanking{} applied to Yosemite, we derive its reduction and conditional preference-risk minimizer, distinguishing label-based classification supervision from cross-input ranking.


To address spatial heterogeneity over the California domain, we condition fire risk on terrain, learned ecoregion embeddings, and past fire rates. These inputs represent spatially varying predictor--fire relationships and provide context for historical retrieval. We restrict pairwise ranking to same-day pairs, because correct ranking across days may result from day-to-day variation rather than correctly resolved spatial differences. We evaluate domain-wide prioritization under a fixed daily budget and, separately, recall when dry matter (DM) consumption exceeds certain thresholds. These diagnostics measure budgeted occurrence ranking and sensitivity to high observed activity.


Our contributions are:
\begin{itemize}
    \item \textbf{A unified adaptation framework.} \method{} combines historical risk pretraining, target-conditioned retrieval, and ranking adaptation without target labels.
    \item \textbf{A budgeted-ranking certificate and selective target.} 
    We show that same-day pairwise losses with nonnegative targets, including direct and selective ranking, bound the number of avoidable misses under a fixed daily budget.
    \item \textbf{Daily prioritization over a heterogeneous grid.} Spatial context and same-day ranking support temporal adaptation across historically observed California locations.
    \item \textbf{Budgeted ranking and high-DM recall.} Ranking adaptation outperforms continued focal fine-tuning: at the same daily budget, selective ranking captures about 344 additional positive location--days (after neighborhood smoothing). All three ranking objectives also improve recall of high-DM fires across severity thresholds.
\end{itemize}

Experimental sections and appendices detail each evaluation protocol.

\section{Problem setup}
\label{sec:problem}


Let $x$ denote input features for a location--day and $y\in\{0,1\}$ indicate whether the neighborhood-smoothed estimate of burned dry matter is positive.  Historical labeled examples form
$\dtrain=\{(x_i,y_i)\}_{i=1}^{n}$.  At adaptation time, the learner receives an unlabeled \emph{transductive} batch $\xtarget=\{x_j^t\}_{j=1}^{m}$ from a later year; it never trains on target-year labels.


A classifier produces a scalar logit $z_\theta(x)$ with $p_\theta(y=1\mid x)=\sigma(z_\theta(x))$.  Historical risk pretraining produces parameters $\theta_{\mathrm{ref}}$ and logit $r(x)$, which provide the starting point for adaptation. We call this model the pretrained model and, when its frozen scores define loss targets, the reference model. Every ranking variant initializes at $\theta_{\mathrm{start}}=\theta_{\mathrm{ref}}$ and fine-tunes the inherited representation on retrieved historical labels. We retain the subscript $\mathrm{ref}$ for the pretrained checkpoint; RDPO and selective ranking also use its frozen scores to define their loss targets.  We evaluate \textbf{ranking} (AUROC, noninterpolated average precision (AP; reported as AUPRC in Yosemite)), \textbf{detection at a fixed rule} (recall, F1 using a historical F1-selected threshold), and \textbf{top-ranked event recall} (pooled top-20\%/30\% recall in Yosemite, daily Top5\% recall in California).

\paragraph{Fire-prone location prioritization under a daily budget.}
For date $t$, let $\eta_{t,i}=\Pr(Y_{t,i}=1\mid x_{t,i})$ denote the fire propensity of candidate cell $i$. Under budget $B$, the policy maximizing the expected number of captured fire-positive cells is $S_t^*=\arg\max_{|S|\leq B}\sum_{i\in S}\eta_{t,i}=\operatorname{TopB}_i\eta_{t,i}$. The model substitutes its learned score for the unknown $\eta_{t,i}$. In California, $B=\max\{1,\lceil0.05|\mathcal{C}_t|\rceil\}=34$ of 666 cells per day; daily AP, recall, and NDCG assess this within-date ordering on fire dates. Pairs are drawn across days in the Yosemite case and within the same day in the California case (Section~\ref{sec:spatial-extension}).

\section{Related work}
\label{sec:related}

\paragraph{Data-driven wildfire prediction.}
Data-driven wildfire prediction complements process-based models \citep{jain_2020,reichstein2019deep}, using reanalysis and remote sensing inputs \citep{abatzoglou2013gridmet,vanderwerf2025gfed5} to model fire risk at regional to continental scales \citep{huot2022next,kondylatos_2022,prapas2023deep}. Our work studies how historically pretrained risk models can be adapted through retrieval and pairwise fine-tuning, with temporal evaluations of the Yosemite case and spatial prioritization across the domain of California.

\paragraph{Long-tailed and cost-sensitive learning.}
Standard remedies for severely imbalanced wildfire labels \citep{he2009learning} include oversampling \citep{chawla2002smote}, focal loss and class-balanced reweighting \citep{lin2017focal,cui2019class}, logit adjustment \citep{menon2021longtail,cao2019ldam}, and decoupled classifiers \citep{kang2020decoupling}. Cost-sensitive learning formalizes asymmetric misclassification costs \citep{elkan2001foundations}. Proposition~\ref{prop:minimizer} connects the \dporanking{} preference-risk minimizer to cost-sensitive logit adjustment; a weighted-focal control tests static reweighting.

\paragraph{Adaptation under distribution shift.}
Importance weighting \citep{shimodaira2000improving,sugiyama2012covariate}, domain-adaptation theory \citep{bendavid2010theory}, adversarial domain alignment \citep{ganin2016domain}, and distributionally robust optimization \citep{sagawa2020distributionally,koh2021wilds} address source--target mismatch. Test-time methods adapt with unlabeled target data \citep{sun2020ttt,wang2021tent}. Our approach instead retrieves labeled historical neighbors \citep{cover1967nearest,khandelwal2020generalization,lewis2020retrieval} conditioned on the target batch, making it batch-transductive \citep{vapnik1998statistical} rather than source-only.

\paragraph{Preference optimization and pairwise ranking.}
DPO reparameterizes preference optimization as a closed-form loss \citep{rafailov2023dpo}, with extensions to alternative divergences \citep{azar2024general} and unpaired objectives \citep{ethayarajh2024kto}. Pairwise learning to rank---RankNet \citep{burges2005ranknet}, LambdaMART \citep{burges2010ranknet}, listwise \citep{cao2007listnet}, and SVM-based \citep{joachims2002optimizing} approaches (see \citealt{liu2009learning} for a survey)---optimizes inter-example order. RDPO compares reference-relative preferences over candidates. Its binary-label specialization, \dporanking{} \citep{jiang2026eapo}, compares labels for one input and reduces to a reference-centered classification surrogate; cross-input RDPO instead compares positive and negative candidate inputs. Direct, RDPO, and selective ranking are alternative objectives for adapting the same pretrained risk model.

\section{Method}
\label{sec:method}

Given a labeled historical archive $\dtrain$ and an unlabeled target batch $\xtarget$, \method{} produces an adapted scoring model through \emph{Pretrain--Retrieve--Rank}. The stages separate representation learning, target-conditioned sample selection, and decision-aligned adaptation, while keeping target labels withheld until evaluation.

\subsection{Pretrain--Retrieve--Rank}

\paragraph{Pretrain.}
Historical focal training learns a risk representation with parameters $\theta_{\mathrm{ref}}$ and score $r(x)$ from labeled source years, before any target-year outcomes are observed. We copy this checkpoint and fine-tune the full representation rather than fitting a new model from the smaller retrieved set. The checkpoint initializes every adaptation policy, while its frozen, evaluation-mode scores also define the targets used by RDPO and selective ranking. Reference scores are detached, so gradients update only the adapted policy. Direct ranking shares the same initialization but does not use reference scores in its loss, giving all variants a controlled common starting point.

\paragraph{Retrieve.}
Continuous covariates are standardized using historical training statistics. Each unlabeled target input then retrieves its $k$ nearest labeled historical inputs in covariate space. This uses target covariates to identify the relevant part of the historical distribution without using target outcomes. The local adaptation set is the union of the retrieved examples,
\begin{equation}
    \local = \bigcup_{x^t\in\xtarget}
    \left\{(x_i,y_i)\in\dtrain: x_i\in\operatorname{KNN}_k(x^t;\dtrain)\right\}.
    \label{eq:retrieval}
\end{equation}
Taking a union deduplicates historical rows retrieved by multiple target inputs, preventing target density from implicitly reweighting the loss. The resulting set is target-conditioned but historically supervised. Let $\extreme=\{(x,y)\in\local:y=1\}$ and $\pi_+=|\extreme|/|\local|$. We fix $k=5$ throughout repeated and rolling evaluations. Retrieval uses weather covariates in the Yosemite setting and also incorporates spatial context in the California setting (Section~\ref{sec:spatial-extension}). This batch-transductive step uses the full set of target covariates only to construct $\local$.

Retrieval performs sample selection rather than label propagation: a target input does not inherit a neighbor's label. The target batch instead identifies a local historical slice for adaptation under covariate shift. In rolling evaluation, the archive, standardization statistics, and causal history features use only information available before the target year. The method is therefore transductive in target covariates but label-free for the target period.

This design assumes useful overlap between historical and target covariates. KNN always returns neighbors even when their absolute distances are large, so retrieval cannot reconstruct a target regime absent from the archive. Starting from the global pretrained representation reduces reliance on the smaller local subset, while held-out rolling years test whether the locality assumption transfers. We keep $k$ and the distance representation fixed across evaluation years rather than tuning them on target outcomes.

\paragraph{Rank.}
Historical labels convert $\local$ into positive--negative pairs $(x^+,x^-)$. The comparisons penalize event--non-event inversions and are invariant to a common score shift, aligning training with AP and budgeted recall. Pair construction follows the downstream decision: Yosemite's direct and selective variants compare inputs across the retrieved set, whereas California pairs cells within a date to prioritize the most fire-prone locations under a daily monitoring budget. Write $s_\theta=z_\theta$ for the adapted logit and define the adapted and frozen pretrained score gaps
\begin{equation}
\Delta_\theta=s_\theta(x^+)-s_\theta(x^-),\qquad
\Delta_{\mathrm{ref}}=r(x^+)-r(x^-).
\label{eq:ranking-gaps}
\end{equation}
For a detached target $\tau$, the per-pair loss is $\ell_\tau(\Delta_\theta)=\operatorname{softplus}[-\beta_{\mathrm{rank}}(\Delta_\theta-\tau)]/\beta_{\mathrm{rank}}$, with $\beta_{\mathrm{rank}}>0$. Direct ranking sets $\tau=0$ and rewards any increase of the positive score relative to the negative \citep{burges2005ranknet}; RDPO and selective ranking instead derive $\tau$ from the pretrained gap (Section~\ref{sec:unified-ranking}). Minibatches sample admissible pairs while adaptation updates the full predictor, including learned embeddings. Afterward, the retrieved set and frozen reference are discarded; inference uses one forward pass through the adapted model, with no nearest-neighbor query or second deployed model.

\subsection{Three adaptation targets and gradient analysis}
\label{sec:unified-ranking}

The three objectives share initialization and pair supervision, with different target gaps. Setting $\tau=\alpha\Delta_{\mathrm{ref}}$ for $\alpha\in[0,1]$ gives
\begin{equation}
\mathcal{L}_{\alpha}
=-\frac{1}{\beta_{\mathrm{rank}}}\mathbb{E}_{(x^+,x^-)}
\log\sigma\!\left[\beta_{\mathrm{rank}}(\Delta_\theta-\alpha\Delta_{\mathrm{ref}})\right].
\label{eq:unified-ranking}
\end{equation}
At $\alpha=0$, direct ranking optimizes the adapted gap; at $\alpha=1$, RDPO optimizes its change from the pretrained gap \citep{rafailov2023dpo}. Under $\pi_\theta(i\mid\mathcal{C})\propto\exp s_\theta(x_i)$, the chosen--rejected log-ratio difference is $\Delta_\theta-\Delta_{\mathrm{ref}}$. Intermediate $\alpha$ retains part of the reference margin. Yosemite applies the same construction to binary labels; Appendix~\ref{app:binary-label-objective} gives its reduction and conditional optimum.

\paragraph{Selective target.}
\label{sec:selective-ranking}

Selective ranking assigns a zero target to pretrained misordered pairs and retains the pretrained gap for correctly ordered pairs. Setting $\alpha(x^+,x^-)=\mathbf{1}\{\Delta_{\mathrm{ref}}>0\}$ yields
\begin{equation}
\mathcal{L}_{\mathrm{sel}}
=-\frac{1}{\beta_{\mathrm{rank}}}\mathbb{E}
\log\sigma\!\left[\beta_{\mathrm{rank}}\bigl(\Delta_\theta-[\Delta_{\mathrm{ref}}]_+\bigr)\right],
\quad [u]_+=\max(0,u).
\label{eq:selective-ranking}
\end{equation}
It equals direct ranking on pretrained errors and RDPO on correctly ordered pairs. The reference is detached for both reference-dependent targets. These soft losses favor increasing the gap; shared parameters, finite training, and regularization do not guarantee correction or retention of every pair. Proposition~\ref{prop:daily-budget-bound} gives both direct and selective ranking a finite-sample certificate for avoidable Top-$B$ misses. Selective ranking additionally provides a positive-reference-margin certificate. California evaluates these objectives at the same daily budget, where all three improve mean recall over continued focal (Section~\ref{sec:california-results}).

\subsubsection{Gradient allocation across pair orderings}

Differentiating the per-pair loss with respect to the adapted gap reveals how each target allocates training pressure to different pair orderings.
\begin{proposition}[Pairwise target gradient allocation]
\label{prop:ranking-gradient}
For a fixed pair and detached reference, the magnitude of the negative derivative of the per-pair loss with respect to $\Delta_\theta$ is
\begin{equation}
g_\alpha=\sigma\!\left[-\beta_{\mathrm{rank}}(\Delta_\theta-\alpha\Delta_{\mathrm{ref}})\right].
\end{equation}
If initial policy and reference scores coincide, then
\begin{equation}
g_\alpha^{(0)}=\sigma\!\left[-\beta_{\mathrm{rank}}(1-\alpha)\Delta_{\mathrm{ref}}\right],
\qquad
\frac{\partial g_\alpha^{(0)}}{\partial\alpha}
=\beta_{\mathrm{rank}}\Delta_{\mathrm{ref}}g_\alpha^{(0)}(1-g_\alpha^{(0)}).
\label{eq:ranking-initial-gradient}
\end{equation}
Thus decreasing $\alpha$ increases the initial gap-gradient magnitude on reference-wrong pairs ($\Delta_{\mathrm{ref}}<0$), while increasing $\alpha$ increases it on reference-correct pairs ($\Delta_{\mathrm{ref}}>0$). At $\alpha=1$ it equals $1/2$ for every pair.
\end{proposition}

The initialization identity assumes equal scores. Policy training-mode and pretrained-model evaluation-mode batch normalization can break this equality; Appendix~\ref{app:alpha-collapse} checks realized scalar weights. The proposition concerns gap gradients, with parameter-gradient scope detailed in Appendix~\ref{app:theory}.

Under score-equal initialization, selective ranking has gap-gradient magnitude $\max\{\sigma(-\beta_{\mathrm{rank}}\Delta_{\mathrm{ref}}),1/2\}$: it preserves direct ranking's larger weight on pretrained errors and RDPO's weight on correctly ordered pairs (Appendix Figure~\ref{fig:ranking-gradients}).

\paragraph{When the targets differ.}
The scaled reference gap controls how strongly the choice of target changes the scalar gradient weight.
\begin{proposition}[Pairwise target sensitivity bound]
\label{prop:alpha-collapse}
For a fixed pair and detached reference, $g_\alpha$ is Lipschitz in $\alpha$ with
\begin{equation}
\left|g_\alpha-g_{\alpha'}\right|
\le\frac{\beta_{\mathrm{rank}}\left|\Delta_{\mathrm{ref}}\right|}{4}\left|\alpha-\alpha'\right|,
\qquad \alpha,\alpha'\in[0,1],
\label{eq:alpha-collapse}
\end{equation}
and at score-equal initialization $g_0^{(0)}-g_1^{(0)}=\sigma(-\beta_{\mathrm{rank}}\Delta_{\mathrm{ref}})-\tfrac12=-\beta_{\mathrm{rank}}\Delta_{\mathrm{ref}}/4+O((\beta_{\mathrm{rank}}\Delta_{\mathrm{ref}})^3)$. For the selective target,
\begin{equation}
\mathbb{E}\bigl|g_{\mathrm{sel}}^{(0)}-g_1^{(0)}\bigr|
\le\frac{\beta_{\mathrm{rank}}}{4}\,
\mathbb{E}\!\left[\left|\Delta_{\mathrm{ref}}\right|\mathbf{1}\{\Delta_{\mathrm{ref}}<0\}\right].
\label{eq:selective-attenuation}
\end{equation}
\end{proposition}

Small $\beta_{\mathrm{rank}}|\Delta_{\mathrm{ref}}|$ makes the weights close; larger gaps allow stronger separation. Selective ranking departs from RDPO only on reference-wrong pairs. Exceeding a negative RDPO target can leave $\Delta_\theta<0$, whereas direct and selective ranking use zero targets on those pairs. These are soft objectives, so task-level correction and retention require empirical evaluation. Appendix~\ref{app:alpha-collapse} provides the proof and measured margin scales.

\subsection{Daily risk prioritization under spatial heterogeneity}
\label{sec:spatial-extension}

The gradient allocation analysis characterizes how each ranking target weights pairwise updates; applying these objectives to a spatially heterogeneous system also requires specifying how locations are represented and compared. In California, we adapt the Pretrain--Retrieve--Rank framework by incorporating location-dependent context into risk prediction and retrieval, and by constructing same-day spatial pairs that align adaptation with daily statewide prioritization.

\subsubsection{Location-dependent risk representation}

We apply these objectives to domain-wide spatial risk prioritization. Let $i$ index a cell, $t$ a date, $w_{t-1,i}$ its previous-day weather, $a_i$ its coordinates and terrain, $e_i$ its ecological region, and $\mathcal{H}_{<\operatorname{yr}(t),i}$ its outcomes from strictly earlier calendar years. The conditional risk is
\begin{equation}
\eta_{t,i}=\Pr\!\left(Y_{t,i}=1\mid w_{t-1,i},a_i,e_i,
\mathcal{H}_{<\operatorname{yr}(t),i},d_t\right),
\label{eq:spatial-risk}
\end{equation}
where $d_t$ encodes seasonality and $\operatorname{yr}(t)$ denotes the calendar year. California pretrains a shared model on
\begin{equation}
u_{t,i}=\left[w_{t-1,i},\;a_i,\;h_{\operatorname{yr}(t),i},\;d_t,\;e_i\right],
\label{eq:spatial-input}
\end{equation}
where $h$ summarizes cell fire rates from strictly earlier years and $e_i$ is the raw EPA Level III category \citep{epa_ecoregions,omernik2014ecoregions}. With $c_{t,i}=[w_{t-1,i},a_i,h_{\operatorname{yr}(t),i},d_t]$, the pretrained and adapted models use learned four-dimensional embeddings:
\begin{equation}
r(u_{t,i})=f_{\mathrm{ref}}([c_{t,i},E_{\mathrm{ref}}(e_i)]),\qquad
s_\theta(u_{t,i})=f_\theta([c_{t,i},E_\theta(e_i)]).
\label{eq:spatial-models}
\end{equation}
This shared representation permits location-dependent baseline risks and weather--landscape interactions. The adapted embedding starts from its pretrained counterpart and is fine-tuned with the full model. The 26 raw fields include weather, terrain, cell history, seasonality, and one region category. History summaries use full historical label grids from strictly earlier years; target-year outcomes never enter them.

\paragraph{Retrieval with weather and spatial context.}

California retrieval uses standardized continuous covariates, excluding history length, plus a fixed ecoregion one-hot vector in Euclidean distance. This representation is separate from the predictor's learned embedding and allows borrowing across regions without restricting neighbors to the same region. Equation~\ref{eq:retrieval} forms the deduplicated adaptation set with $k=5$. Appendix~\ref{app:california} defines the inputs, causal-history shrinkage, and exact distance.

\subsubsection{Same-day spatial comparisons}

We construct positive--negative pairs from the same historical day because cross-date ranking can reward separating high-risk from low-risk dates without resolving which locations are riskier within a date. We therefore construct. To see the cancellation, write $s_\theta(u_{t,i})=b_\theta(t)+q_\theta(t,i)$, where $b_\theta(t)$ is any additive date-specific offset. Then
\begin{equation}
s_\theta(u_{t,i^+})-s_\theta(u_{t,i^-})
=q_\theta(t,i^+)-q_\theta(t,i^-).
\label{eq:same-day-cancellation}
\end{equation}
The ranking term therefore cannot improve by raising all scores on a high-risk date: it must distinguish positive from negative cells within that date. Both within-region and cross-region comparisons are allowed.

For a minibatch $B\subset\local$, let $\mathcal{T}_B$ contain its dates with both labels. For each $t\in\mathcal{T}_B$, sample one positive cell $i_t^+$ and one negative cell $i_t^-$ uniformly within their classes. Set $\Delta_{\theta,t}=s_\theta(u_{t,i_t^+})-s_\theta(u_{t,i_t^-})$, with $\Delta_{\mathrm{ref},t}$ defined analogously. The pairwise term is
\begin{equation}
\mathcal{L}_{\mathrm{pair}}(B;\tau)
=\frac{1}{|\mathcal{T}_B|}\sum_{t\in\mathcal{T}_B}
\frac{\operatorname{softplus}\!\left[-\beta_{\mathrm{rank}}
(\Delta_{\theta,t}-\tau_t)\right]}{\beta_{\mathrm{rank}}},
\label{eq:spatial-pair}
\end{equation}
with zero contribution when $\mathcal{T}_B$ is empty. The targets analyzed in Section~\ref{sec:unified-ranking} are
\begin{equation}
\tau_t^{\mathrm{direct}}=0,\qquad
\tau_t^{\mathrm{RDPO}}=\Delta_{\mathrm{ref},t},\qquad
\tau_t^{\mathrm{selective}}=[\Delta_{\mathrm{ref},t}]_+,
\label{eq:spatial-targets}
\end{equation}
where $[u]_+=\max(0,u)$. RDPO ranks the implicit residual $h_\theta^{\mathrm{res}}(u)=s_\theta(u)-r(u)$. California fine-tunes with the pairwise objective, without focal supervision:
\begin{equation}
\mathcal{L}_{\mathrm{CA}}=
\mathbb{E}_{B,\,\mathrm{pair\ draws}}\!\left[\lambda_{\mathrm{rank}}\mathcal{L}_{\mathrm{pair}}(B;\tau)
+\lambda_s M_{\mathrm{loss}}(B)\right],
\label{eq:california-ranking}
\end{equation}
where $M_{\mathrm{loss}}$ is the TabNet mask term. We use $\beta_{\mathrm{rank}}=\lambda_{\mathrm{rank}}=0.1$ and $\lambda_s=0.05$; the single all-feature attention group makes the mask term zero. Focal loss is detached and monitored only during ranking adaptation. Dates without both classes contribute no preference pair. No positive-only preference term or margin cap is used. Both California and Yosemite exclude focal supervision from ranking adaptation; California uses same-day spatial pairs, whereas Yosemite's direct and selective variants use cross-date pairs at $\beta_{\mathrm{rank}}=1$. Appendix~\ref{app:california} details optimization and evaluation, and Appendix~\ref{app:broader-design} describes broader-area extensions.

\section{Yosemite: temporal shift}
\label{sec:experiments}

\paragraph{Setup.}

For the Yosemite case, we evaluate \emph{temporal shift} on a fixed 3$\times$3 grid. The same nine cells appear in historical and target years, holding the candidate geography fixed while assessing year-to-year adaptation. We use daily GridMET meteorology \citep{abatzoglou2013gridmet} and GFED-derived dry matter \citep{vanderwerf2025gfed5} with twelve weather inputs. The main split contains 62,460 training grid--days through 2020 (7,613 positives) and 3,285 target grid--days in 2021 (363 positives). The historically pretrained TabNet \citep{arik2021tabnet} is adapted for 100 epochs with AdamW at $10^{-4}$; retrieval uses $k=5$. The adapted local policies share the same retrieved rows, frozen pretrained model, initialization, and seeded batches. The main seeds are $\{11,17,23,29,37\}$; target labels are used only for evaluation. Full architecture and optimization details appear in Appendix~\ref{app:implementation}. We additionally train \textbf{joint focal + direct ranking} from random initialization on all historical rows, using the same 50-epoch Adam protocol as the pretrained focal model and a direct-ranking coefficient of one. This from-scratch baseline does not use target-dependent retrieval; its full configuration is in Appendix~\ref{app:joint-training}.

The six local comparison policies are: the \textbf{frozen pretrained model} (no local adaptation), \textbf{continued focal} (local focal fine-tuning), \textbf{weighted focal} (Equation~\ref{eq:coefficient-matched}), \textbf{direct pairwise ranking} (Equation~\ref{eq:unified-ranking} at $\alpha=0$), \textbf{selective pairwise ranking} (Equation~\ref{eq:selective-ranking}), and \textbf{\dporanking{}} (Equation~\ref{eq:eapo}). All three ranking adaptations fix $\beta=1$ and preference coefficient one, with no focal supervision or extra positive-label term ($\lambda_2=0$). Direct and selective ranking compare positive--negative inputs across dates; binary-label RDPO compares true and complementary labels for each input. These configurations share training settings while differing in candidate definitions. Bold table entries indicate the best mean per column. Expanding historical windows end in 2018, 2019, and 2020 for rolling tests in 2019, 2020, and 2021, with seeds $\{11,17,23\}$. Uncertainty intervals, metric definitions, and evaluation scope are in Appendix~\ref{app:implementation}.

\paragraph{Results.}
\label{sec:results}

Pairwise fine-tuning improves pooled ranking on the 2021 holdout (Table~\ref{tab:main-2021}). Direct ranking raises AUROC from 70.32\% for the pretrained model to 71.23\%, and AUPRC from 21.23\% to 22.82\%. Selective ranking attains AUROC of 71.51\% and AUPRC of 22.94\%. The pointwise \dporanking{} baseline also improves over the pretrained model, reaching AUROC of 71.00\% and AUPRC of 22.59\%. These results show that the inherited risk representation supports several subsequent adaptation objectives.

\begin{table}[t]
\centering
\caption{Results for Yosemite in 2021, showing mean $\pm$ SD (\%) over five seeds, with the highest mean in each column in bold. Ranking adaptation fixes $\beta=1$ without focal supervision. R@20 and R@30 pool all target location--days.}
\label{tab:main-2021}
\begingroup
\papertablefont
\setlength{\tabcolsep}{2pt}
\renewcommand{\arraystretch}{1.12}
\renewcommand{\tabularxcolumn}[1]{m{#1}}
\begin{tabularx}{\linewidth}{@{}>{\raggedright\arraybackslash}m{70pt}>{\centering\arraybackslash}X>{\centering\arraybackslash}X>{\centering\arraybackslash}X>{\centering\arraybackslash}X>{\centering\arraybackslash}X>{\centering\arraybackslash}X@{}}
\toprule
\textbf{Method} & \textbf{Recall} & \textbf{F1} & \textbf{AUROC} & \textbf{AUPRC} & \textbf{R@20} & \textbf{R@30} \\
\midrule
Pretrained & 57.80\,$\pm$\,5.70 & 30.61\,$\pm$\,0.75 & 70.32\,$\pm$\,0.51 & 21.23\,$\pm$\,1.27 & 42.31\,$\pm$\,1.15 & 57.08\,$\pm$\,2.54 \\\addlinespace[2.5pt]
Joint focal + direct & 52.73\,$\pm$\,6.77 & 29.39\,$\pm$\,1.77 & 69.40\,$\pm$\,1.00 & 20.27\,$\pm$\,0.77 & 40.39\,$\pm$\,2.85 & 54.82\,$\pm$\,3.19 \\\addlinespace[2.5pt]
Continued focal & 57.02\,$\pm$\,6.80 & 29.93\,$\pm$\,1.91 & 70.16\,$\pm$\,1.11 & 21.07\,$\pm$\,1.62 & 41.60\,$\pm$\,2.71 & 56.31\,$\pm$\,3.63 \\\addlinespace[2.5pt]
Weighted focal & 56.09\,$\pm$\,6.07 & 29.94\,$\pm$\,1.56 & 70.02\,$\pm$\,1.13 & 21.00\,$\pm$\,1.53 & 41.10\,$\pm$\,2.89 & 56.03\,$\pm$\,3.72 \\\addlinespace[2.5pt]
Direct ranking & 60.50\,$\pm$\,2.37 & \textbf{31.31}\,$\pm$\,1.11 & 71.23\,$\pm$\,0.62 & 22.82\,$\pm$\,1.10 & 44.90\,$\pm$\,1.66 & \textbf{58.68}\,$\pm$\,2.15 \\\addlinespace[2.5pt]
binary-label RDPO & 61.27\,$\pm$\,5.86 & 30.79\,$\pm$\,1.41 & 71.00\,$\pm$\,0.81 & 22.59\,$\pm$\,1.70 & 43.86\,$\pm$\,2.62 & 57.36\,$\pm$\,3.25 \\\addlinespace[2.5pt]
Selective ranking & \textbf{61.32}\,$\pm$\,3.50 & 31.25\,$\pm$\,0.86 & \textbf{71.51}\,$\pm$\,0.44 & \textbf{22.94}\,$\pm$\,1.09 & \textbf{45.56}\,$\pm$\,1.37 & \textbf{58.68}\,$\pm$\,1.87 \\
\bottomrule
\end{tabularx}
\endgroup
\end{table}

Rolling evaluation extends the comparison across 2019--2021. Figure~\ref{fig:rolling-ranking} reports paired gains over each pretrained checkpoint across nine year--seed cases, and Appendix Table~\ref{tab:rolling-overall} gives absolute scores. Direct ranking attains mean AUROC of 74.19\% and AUPRC of 24.30\%, compared with 73.12\% and 22.85\% for the pretrained model and 73.08\% and 22.08\% for continued focal. Relative to \dporanking{}, direct ranking gains 2.06 AUROC, 2.25 AUPRC, 3.69 R@20, and 5.97 R@30 percentage points (differences computed before rounding). This supports explicit pairwise fine-tuning as an effective adaptation option under temporal shift. The binary-label RDPO configuration evaluated here, with no focal supervision and a fixed training schedule, shows unstable performance across test years; this observation is specific to the evaluated configuration and does not establish a general limitation of RDPO.

\begin{figure}[t]
\centering
\includegraphics[width=0.9\linewidth]{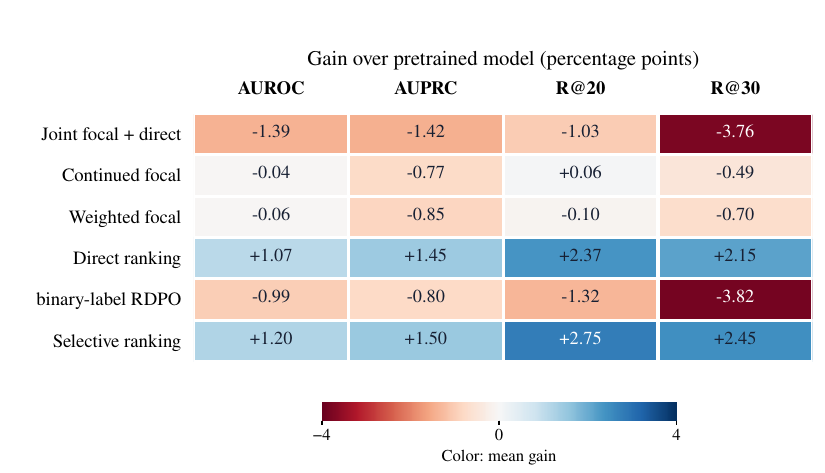}
\caption{Paired Yosemite ranking gains over the pretrained model, 2019--2021, in percentage points. Cells and color show mean gains over three per-seed averages across years. Joint focal + direct starts from random initialization on all historical rows; local adaptation starts from the pretrained focal model. Binary-label RDPO is the pointwise preference baseline, with $\beta=1$ and no focal term. See Appendix Table~\ref{tab:rolling-overall} for absolute scores.}
\label{fig:rolling-ranking}
\end{figure}

The from-scratch joint focal and direct-ranking baseline attains rolling AUROC of 71.73\% and AUPRC of 21.43\%, compared with 74.19\% and 24.30\% for local direct adaptation. The paired local-minus-joint AUROC gain is 2.46 points; the AUPRC gain is 2.86 points. This comparison evaluates the complete local-adaptation configuration, with differences in data access and training schedules documented in Appendix~\ref{app:joint-training}.

On Yosemite's nine-cell grid, the daily Top5\% rule selects one cell, so this diagnostic operates at a different decision scale from statewide monitoring. Detailed direct--selective comparisons, annual results, operating points, and same-day pair diagnostics appear in Appendix~\ref{app:results}. California evaluates adaptation with same-day training pairs on the complete statewide candidate grid.

\section{California: statewide risk prioritization}
\label{sec:california-results}

\paragraph{Complete-grid evaluation.}

California has diverse terrain, ecology, and fire regimes. This domain is discretized into 666 grid cells (0.25$^\circ$). Historical risk pretraining uses 2002--2021 data, and adaptation targets 2022. The fixed 20\% historical sample contains 972,892 cell--days. Batch-transductive $k=5$ retrieval queries all 243,090 unlabeled target covariates, yielding 363,643 unique historical examples for fine-tuning. Evaluation covers the complete daily grid, with 14,760 positives across 351 fire dates. The daily Top5\% rule selects 34 cells per date, or 12,410 alarms in the 2022 evaluation. Within each of seeds \{11, 22, 33\}, adapted models share the pretrained checkpoint, scaler, and retrieved examples; pairwise variants use $\lambda_{\rm rank}=\beta_{\rm rank}=0.1$ and the same pair-sampling rule, with focal supervision excluded from all three ranking objectives. See Appendix~\ref{app:california} for the full protocol.

\begin{table}[t]
\centering
\caption{Results for the entire California domain in 2022, showing Mean $\pm$ SD (\%) over three seeds. Ranking adaptation excludes focal supervision. Daily metrics average fire dates; global metrics pool cell--days.}
\label{tab:california-ranking}
\label{tab:california-fulltarget}
\begingroup
\papertablefont
\setlength{\tabcolsep}{2pt}
\renewcommand{\arraystretch}{1.08}
\begin{tabularx}{\linewidth}{@{}l*{4}{>{\centering\arraybackslash}X}@{}}
\toprule
Method & Daily AP & Daily R@5\% & Global AP & AUROC \\
\midrule
XGBoost & \mbox{14.24\,$\pm$\,0.07} & \mbox{11.64\,$\pm$\,0.23} & \mbox{11.02\,$\pm$\,0.02} & \mbox{66.27\,$\pm$\,0.04} \\
Pretrained spatial focal & \mbox{21.10\,$\pm$\,1.24} & \mbox{17.98\,$\pm$\,1.36} & \mbox{19.30\,$\pm$\,0.63} & \mbox{78.34\,$\pm$\,0.37} \\
Continued focal & \mbox{21.62\,$\pm$\,0.33} & \mbox{18.70\,$\pm$\,0.74} & \mbox{20.07\,$\pm$\,0.34} & \mbox{78.60\,$\pm$\,0.21} \\
Direct ranking & \mbox{24.35\,$\pm$\,0.69} & \mbox{22.25\,$\pm$\,1.39} & \mbox{20.80\,$\pm$\,1.53} & \mbox{79.46\,$\pm$\,0.50} \\
RDPO & \mbox{24.47\,$\pm$\,0.42} & \mbox{22.11\,$\pm$\,1.92} & \mbox{\textbf{20.89}\,$\pm$\,1.39} & \mbox{\textbf{79.47}\,$\pm$\,0.47} \\
Selective ranking & \mbox{\textbf{24.57}\,$\pm$\,0.52} & \mbox{\textbf{22.79}\,$\pm$\,2.07} & \mbox{20.71\,$\pm$\,1.13} & \mbox{79.38\,$\pm$\,0.33} \\
\bottomrule
\end{tabularx}
\endgroup
\end{table}

\paragraph{Location prioritization under a tight daily budget.}

Without focal supervision during adaptation, direct, RDPO, and selective ranking attain daily AP of 24.35\%, 24.47\%, and 24.57\%, respectively, compared with 21.62\% for continued focal fine-tuning (Table~\ref{tab:california-ranking}). Their macro Top5\% recall is 22.25\%, 22.11\%, and 22.79\%, versus 18.70\% for continued focal. These gains show the value of same-day ranking supervision for daily location prioritization. Proposition~\ref{prop:daily-budget-bound} connects the nonnegative-target losses used by direct and selective ranking to a finite-sample certificate for avoidable Top-$B$ misses.

At 34 cells per day, selective captures a mean of 3,210.3 positive cell--days versus 2,866.0 for continued focal: 344 additional captures, or approximately 12\%, at the same number of alarms (Appendix Table~\ref{tab:california-fulltarget-budget}). Direct and RDPO add 311 and 301, respectively. These are neighborhood-smoothed positive cell--days, not independent incidents.

\paragraph{Thresholded recall within high-DM strata.}

Among the 2,193 raw-DM-positive test examples, we evaluate training-threshold recall within the highest-DM 5\%, 10\%, and 20\%. All ranking variants identify a much larger fraction than continued focal: direct gains 38.48/27.27/20.27 points, RDPO 38.18/26.97/19.89, and selective 39.70/28.18/20.50 (Figure~\ref{fig:california-raw-dm-recall-gain}). These gains describe sensitivity under each policy's historical F1-selected threshold. Target-year DM defines retrospective strata only and supplies no severity supervision. Alarm rates are not matched here, so this diagnostic does not establish severity forecasting or equal-budget severe-event discrimination.

\begin{figure}[t]
  \centering
  \includegraphics[width=0.68\linewidth]{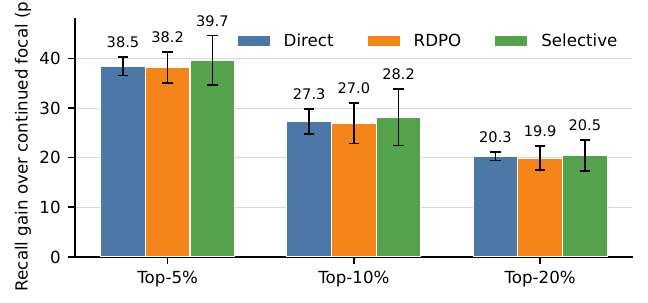}
  \caption{California recall gain over continued focal within retrospective high-DM strata. Mean and sample SD of paired seed differences; each policy uses its historical F1-selected threshold, with unmatched alarm rates.}
  \label{fig:california-raw-dm-recall-gain}
\end{figure}

\paragraph{Isolating retrieval.} A separate three-seed continued-focal ablation compares spatial KNN, a size-matched random subset, and the full historical sample at exactly 6,800 updates. KNN improves mean daily AP by 0.66 points over random and 0.36 over full-history adaptation, with positive paired differences in every seed. This isolates source selection under a common loss and compute budget; gains in Top5\% recall remain less conclusive (Appendix Table~\ref{tab:california-focal-retrieval-multiseed}).


\section{Conclusion}
\label{sec:conclusion}

\method{} combines historical risk pretraining, target-conditioned retrieval, and ranking adaptation without focal supervision. Over the entire domain of California, all three ranking objectives improve mean daily AP and Top5\% recall over continued focal. At a daily budget of 34 locations, the selective configuration captures approximately 344 additional neighborhood-smoothed positive cell--days. Ranking also increases recall within retrospective high-DM strata at training-selected thresholds. 




\section*{AI Use Statement}

In this work, we used generative AI tools to help develop the theoretical framework, formulate mathematical claims, provide critical ingredients for proving mathematical claims, assist in the writing of proofs, provide feedback on experimental design, implement methods, clean and reformat dataset.
We did \emph{not} use generative AI tools to generate synthetic data sets, propose or refine hypotheses, assist with translation, support qualitative and thematic data analysis, or interpret results.
Additionally, we used generative AI tools to help design an overview of the framework (Figure~\ref{fig:framework}), create Figures~\ref{fig:rolling-ranking} and \ref{fig:california-raw-dm-recall-gain} from experimental results, create and edit software code, summarize and analyze existing literature on wildfire research, draft the paper based on results,  edit the paper to improve readability, and format references.
The authors have reviewed all AI-assisted work, including the theoretical claims, experimental design and implementation, software code, and results.
The authors take full responsibility for the final content of this work, including text, claims or artifacts produced with the aid of generative AI.

\section*{Ethics Statement}

This work involves no human subjects, no personally identifying information, and no new dataset release; all inputs are public environmental products \citep{abatzoglou2013gridmet,vanderwerf2025gfed5,epa_ecoregions} whose licenses permit research use. Because risk ranking directs limited monitoring capacity, both error modes are costly---a missed positive can leave a developing fire unobserved, while a false alarm consumes capacity another location needed---so we report ranking, fixed-threshold detection, and top-ranked event recall separately (Section~\ref{sec:problem}) rather than a single headline number. Two risks deserve mention: location prioritization can disadvantage regions with sparse historical support, since our history features shrink such cells toward a statewide prior, so deployments should audit performance by region rather than only in aggregate; and our label is a neighborhood-smoothed occurrence indicator rather than a verified ignition, so predictions should not be read as ignition forecasts. We therefore claim no operational readiness, and regard premature operational use---without the deployment-matched evaluation of Appendix~\ref{app:broader-design}---as the principal misuse risk of this work.

\section*{Reproducibility Statement}

Theoretical claims are stated in Section~\ref{sec:method} and proved in Appendix~\ref{app:theory}, which also gives the expected-batch coefficient qualification and conditions for order preservation. Yosemite data, architecture, optimizer settings, retrieval, policy definitions, seeds, and the rolling protocol appear in Section~\ref{sec:experiments} and Appendix~\ref{app:implementation}, including the fixed-configuration table and initial gradient-norm checks. The California protocol is in Appendix~\ref{app:california}: label smoothing, the fixed 20\% stratified sample and its data seed, causal history features with shrinkage and cold-start rules, retrieval distance, optimizer settings, stopping rule, XGBoost configuration, and training-only threshold selection; metric definitions, including pooled R@$q$ versus daily Top5\% selection and noninterpolated average precision, accompany the results. The protocol records pretraining epoch caps, fixed loss settings, and the exclusion of focal supervision from ranking adaptation. All datasets are public, and our code and configuration files will be released.

\endgroup
\bibliography{rarr}
\bibliographystyle{eapo_iclr_style}

\clearpage

\clearpage
\appendix
\section{Derivations and ordering conditions}
\label{app:theory}

\subsection{Binary-label RDPO: objective and analytical results}
\label{app:binary-label-objective}

Yosemite applies \dporanking{} to the label pair $(y,1-y)$ for one input. With $s_y=2y-1\in\{-1,+1\}$, its preference loss is
\begin{equation}
\ell_{\mathrm{pref}}(z,r,y)
=-\frac{1}{\beta}\log\sigma\!\left(
\beta\left[
\log\frac{p_\theta(y\mid x)}{p_{\mathrm{ref}}(y\mid x)}
-\log\frac{p_\theta(1-y\mid x)}{p_{\mathrm{ref}}(1-y\mid x)}
\right]\right).
\label{eq:dpo-original}
\end{equation}
The $1/\beta$ scale leaves the isolated preference-risk minimizer unchanged.

With uniform averages over retrieved rows, the evaluated baseline optimizes
\begin{equation}
\mathcal{L}_{\mathrm{binary\text{-}RDPO}}
=\mathbb{E}_{(x,y)\sim\local}
\!\left[\ell_{\mathrm{pref}}(z_\theta(x),r(x),y)\right]
+\lambda_s M_{\mathrm{loss}}.
\label{eq:eapo}
\end{equation}
We fix $\beta=1$ and the preference coefficient to one, with no focal term and no extra positive-label preference term ($\lambda_2=0$). The mask coefficient is $\lambda_s=0.05$; its realized term is zero.

\subsubsection{Binary reduction and conditional minimizer}

\begin{proposition}[Binary reduction]
\label{prop:binary-reduction}
Assume $\beta>0$, $y\in\{0,1\}$, and let $z,r\in\mathbb{R}$ be Bernoulli policy and reference logits, respectively.  Then Equation~\ref{eq:dpo-original} equals
\begin{equation}
    \ell_{\mathrm{pref}}(z,r,y)
    =-\frac{1}{\beta}\log\sigma\!\left(\beta s_y(z-r)\right).
    \label{eq:binary-reduction}
\end{equation}
Its gradient and curvature with respect to $z$ are
\begin{align}
\frac{\partial\ell_{\mathrm{pref}}}{\partial z}
&=-s_y\,\sigma\!\left(-\beta s_y(z-r)\right),\\
\frac{\partial^2\ell_{\mathrm{pref}}}{\partial z^2}
&=\beta\,\sigma\!\left(\beta s_y(z-r)\right)
\sigma\!\left(-\beta s_y(z-r)\right)>0.
\end{align}
\end{proposition}

With label-policy gaps $\Delta_\theta=s_yz_\theta(x)$ and $\Delta_{\mathrm{ref}}=s_yr(x)$, this is Equation~\ref{eq:unified-ranking} at $\alpha=1$. Uniform averaging and the absence of an extra positive-label term give
\begin{equation}
    a_0=a_1=1.
    \label{eq:coefficients}
\end{equation}
Allowing general fixed class coefficients makes the reference's effect on the conditional optimum explicit.
\begin{proposition}[DPO-only population minimizer]
\label{prop:minimizer}
Fix $a_0,a_1>0$ and $\beta>0$. At a point $x$ with local positive probability $0<\eta(x)<1$ and fixed reference $r(x)$, the conditional preference risk $R(z\mid x)=\eta a_1\ell_{\mathrm{pref}}(z,r,1)+(1-\eta)a_0\ell_{\mathrm{pref}}(z,r,0)$ is strictly convex in $z$ with unique minimizer
\begin{equation}
z^*_{\mathrm{pref}}(x)=r(x)+\frac{1}{\beta}
\left[\operatorname{logit}\eta(x)+\log\frac{a_1}{a_0}\right].
\label{eq:minimizer}
\end{equation}
\end{proposition}
For the evaluated setting, $\beta=1$ and $a_0=a_1=1$, so $z^*_{\mathrm{pref}}(x)=r(x)+\operatorname{logit}\eta(x)$. The objective is therefore a pointwise binary-classification preference loss rather than the positive--negative input-ranking loss used for California. It uses no focal term or additional positive-label weight. The reference contributes an input-dependent offset: a reference nondecreasing in $\eta$ preserves ordering, whereas a sufficiently misordered reference can reverse it.

\subsubsection{Weighted focal control}

We retain a static weighted-focal control for cost-sensitive adaptation. It multiplies the aligned focal loss by fixed class weights
\begin{equation}
    c_0=2,\qquad c_1=2+0.1/\pi_+.
    \label{eq:coefficient-matched}
\end{equation}
These weights are fixed using retrieved-set prevalence and are independent of minibatch composition. This is a separate class-reweighting baseline, rather than a coefficient match to the pure binary-label RDPO objective (Appendix~\ref{app:implementation}).

\subsection{Proof of Proposition~\ref{prop:binary-reduction}}

For a Bernoulli model with logit $z$,
\begin{equation}
\log\frac{p_\theta(1\mid x)}{p_\theta(0\mid x)}=z,
\qquad
\log\frac{p_\theta(0\mid x)}{p_\theta(1\mid x)}=-z.
\end{equation}
The same identities hold for the reference logit $r$.  If $y=1$, the expression inside the square brackets of Equation~\ref{eq:dpo-original} is $z-r$; if $y=0$, it is $-(z-r)$.  Both cases equal $s_y(z-r)$, which yields Equation~\ref{eq:binary-reduction}.  Let $u=\beta s_y(z-r)$.  Since $d[-\log\sigma(u)]/du=-\sigma(-u)$, the chain rule gives
\begin{equation}
\frac{\partial\ell_{\mathrm{pref}}}{\partial z}
=-s_y\sigma(-u).
\end{equation}
Differentiating again gives $\beta\sigma(u)\sigma(-u)>0$, where strict positivity follows from $\beta>0$ and finite $z,r$.

\subsection{Proof of Proposition~\ref{prop:minimizer}}

Fix $a_0,a_1>0$ and $\beta>0$.  At a fixed $x$, write $d=z-r$ and $0<\eta=\Pr_{Q_{\mathrm{local}}}(Y=1\mid X=x)<1$.  The conditional preference risk is
\begin{equation}
R(d)=-\frac{1}{\beta}\left[
\eta a_1\log\sigma(\beta d)
+(1-\eta)a_0\log\sigma(-\beta d)
\right].
\end{equation}
Differentiating twice gives
\begin{align}
R'(d)&=-\eta a_1\sigma(-\beta d)
+(1-\eta)a_0\sigma(\beta d),\\
R''(d)&=\beta\left[\eta a_1+(1-\eta)a_0\right]
\sigma(\beta d)\sigma(-\beta d)>0.
\end{align}
Hence $R$ is strictly convex.  Its unique stationary point, and therefore its unique global minimizer, satisfies $R'(d)=0$.
Set $q=\sigma(\beta d)$.  Then
$(1-\eta)a_0q=\eta a_1(1-q)$, so
\begin{equation}
q=\frac{\eta a_1}{\eta a_1+(1-\eta)a_0}.
\end{equation}
Taking odds and solving for $d$ yields
\begin{equation}
d^*=\frac{1}{\beta}\left[\operatorname{logit}\eta+\log(a_1/a_0)\right],
\end{equation}
which proves Equation~\ref{eq:minimizer}.

\subsection{Conditions for order preservation}

For two inputs $x_i,x_j$ with $0<\eta(x_j)<\eta(x_i)<1$ and the same $a_0,a_1>0$ and $\beta>0$,
\begin{equation}
z^*_{\mathrm{pref}}(x_i)-z^*_{\mathrm{pref}}(x_j)
=r(x_i)-r(x_j)
+\frac{1}{\beta}\left[
\operatorname{logit}\eta(x_i)-\operatorname{logit}\eta(x_j)
\right].
\label{eq:pair-difference}
\end{equation}
If $\eta(x_i)>\eta(x_j)$ and $r(x_i)\ge r(x_j)$, the reference difference is nonnegative and the posterior-log-odds difference is strictly positive, so the order is preserved.  If the reference is misordered, a reversal occurs whenever
\begin{equation}
r(x_j)-r(x_i)
>\frac{1}{\beta}
\left[\operatorname{logit}\eta(x_i)-\operatorname{logit}\eta(x_j)\right].
\end{equation}
Equality gives a tie; strict inequality gives a reversal. The condition identifies how the frozen pretrained offset affects the ordering of the pointwise preference-risk minimizer.

\subsection{Full-set versus minibatch normalization}

The pure binary-label RDPO baseline averages its preference term over all rows in each minibatch; both label coefficients remain one. For fixed pointwise scores, an i.i.d. minibatch mean is an unbiased estimate of the retrieved-distribution preference risk. Training-mode batch normalization makes policy scores depend on the batch, so this identity does not establish the exact neural-network training risk. The weighted-focal control instead uses the fixed retrieved-set weights in Equation~\ref{eq:coefficient-matched}.

\subsection{Scope of the conditional minimizer}

Proposition~\ref{prop:minimizer} analyzes the preference component with fixed coefficients and unrestricted deterministic logits. The implemented baseline uses this preference loss without focal supervision or extra positive-label weighting; the TabNet mask term is zero. Shared parameters and batch normalization couple the scores, so the neural-network optimizer trajectory is outside the unrestricted pointwise result.

\subsection{Proof of Proposition~\ref{prop:ranking-gradient}}

For one candidate pair, write $u=\beta_{\mathrm{rank}}(\Delta_\theta-\alpha\Delta_{\mathrm{ref}})$ and hold the reference fixed. The chain rule gives
\begin{equation}
\frac{\partial\ell_\alpha}{\partial\Delta_\theta}=-\sigma(-u),\qquad
\frac{\partial^2\ell_\alpha}{\partial\Delta_\theta^2}
=\beta_{\mathrm{rank}}\sigma(u)\sigma(-u)>0.
\end{equation}
Substituting score-equal initialization, $\Delta_\theta=\Delta_{\mathrm{ref}}$, gives Equation~\ref{eq:ranking-initial-gradient}. Differentiating with respect to $\alpha$ gives the displayed derivative, whose sign equals the reference gap. Therefore the initial direct-ranking gap gradient is greater than $1/2$ on wrong pairs, less than $1/2$ on correct pairs, and equal to $1/2$ on ties; RDPO gives $1/2$ for all pairs. The per-pair parameter gradient is
\begin{equation}
\nabla_\theta\ell_\alpha=-g_\alpha\bigl[\nabla_\theta s_\theta(x^+)-\nabla_\theta s_\theta(x^-)\bigr].
\end{equation}
Different pair Jacobians and cancellation in their average prevent treating the scalar factor as a universal parameter-gradient norm comparison. Likewise, adaptive optimizers need not take proportionally larger steps.

\subsection{Selective target and gradient allocation}

\begin{figure}[t]
\centering
\includegraphics[width=0.62\linewidth]{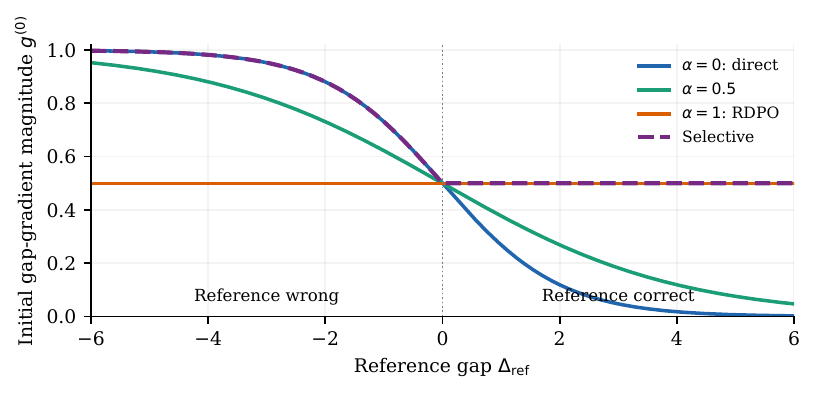}
\caption{Analytic initial gap-gradient magnitudes for Equation~\ref{eq:unified-ranking} and selective ranking, assuming equal initial scores and $\beta_{\mathrm{rank}}=1$. Curves characterize the loss under the stated initialization assumption.}
\label{fig:ranking-gradients}
\end{figure}

For $\Delta_{\mathrm{ref}}\leq0$, $[\Delta_{\mathrm{ref}}]_+=0$ and selective equals direct ranking. For $\Delta_{\mathrm{ref}}>0$, selective equals RDPO. The target has no policy derivative because the reference is detached. Under score-equal initialization its gradient factor is
\begin{equation}
g_{\mathrm{sel}}^{(0)}=
\begin{cases}
\sigma(-\beta_{\mathrm{rank}}\Delta_{\mathrm{ref}}),&\Delta_{\mathrm{ref}}\leq0,\\
1/2,&\Delta_{\mathrm{ref}}>0.
\end{cases}
\end{equation}
Figure~\ref{fig:ranking-gradients} illustrates this analytic allocation. For any of these one-sided per-pair losses, the derivative remains negative at every finite gap. The isolated surrogate favors increasing the gap rather than imposing a hard condition $\Delta_\theta>\tau$. In a shared neural network, finite training, competing pairs, and regularization can prevent crossing zero or retaining a previously correct order. This distinction is visible in the correction/retention diagnostics.

\subsection{Fire-prone location prioritization under a daily budget}
\label{app:daily-budget-bound}

For date $t$, let $P_t$ and $N_t$ be the numbers of positives and negatives, let $T_{\theta,t}^B$ be the positives among the $B$ highest-scored cells, and define the avoidable misses relative to a label oracle as $q_t^B=\min(B,P_t)-T_{\theta,t}^B$. Let $\mathcal{P}_t=\{(i,j):y_{t,i}=1,\ y_{t,j}=0\}$ contain all same-day positive--negative pairs. For pair-dependent targets $\tau^{ij}$, write $\ell_\tau^{ij}=\ell_{\tau^{ij}}(\Delta_\theta^{ij})$. Direct and selective ranking use the nonnegative targets $0$ and $[\Delta_{\mathrm{ref}}^{ij}]_+$, respectively.

\begin{proposition}[Nonnegative-target daily-budget certificate]
\label{prop:daily-budget-bound}
For $P_t>0$ and any targets $\tau^{ij}\geq0$,
\begin{equation}
\frac{q_t^B}{P_t}
\leq
\frac{\beta_{\mathrm{rank}}}{P_t\log 2}
\sum_{(i,j)\in\mathcal{P}_t}\ell_\tau^{ij}.
\label{eq:daily-budget-bound}
\end{equation}
Thus the bound applies to both direct and selective ranking. Selective ranking additionally satisfies
\begin{equation}
\ell_{\mathrm{sel}}^{ij}\leq\frac{\log 2}{\beta_{\mathrm{rank}}}
\quad\Longrightarrow\quad
\Delta_\theta^{ij}\geq[\Delta_{\mathrm{ref}}^{ij}]_+.
\label{eq:selective-margin-certificate}
\end{equation}
\end{proposition}

Each avoidable miss can be matched to a selected negative that outranks an unselected positive, giving $\Delta_\theta^{ij}\leq0$. For any $\tau^{ij}\geq0$, every such inversion has $\ell_\tau^{ij}\geq\log 2/\beta_{\mathrm{rank}}$; summing over the matched pairs proves Equation~\ref{eq:daily-budget-bound}. The second statement follows from monotonicity of softplus. The Top-$B$ certificate therefore applies to both direct and selective ranking. Selective's additional certificate says that sufficiently low loss removes a negative ordering on a pretrained inversion and retains the positive reference margin on a correctly ordered pair. RDPO can meet a negative target without correcting an inversion, whereas direct ranking does not retain a positive reference margin.

\begin{proposition}[Weighting induced by the minibatch sampler]
\label{prop:minibatch-pair-weighting}
Fix batch-independent per-example scores and suppose minibatches are exchangeable among examples sharing a date and label. For dates with $P_tN_t>0$, define the full-pair mean
\begin{equation}
\overline{L}_{\tau,t}
=\frac{1}{P_tN_t}\sum_{(i,j)\in\mathcal{P}_t}\ell_\tau^{ij}
\end{equation}
and let
\begin{equation}
w_t=\mathbb{E}_{\mathcal{B}}\!\left[
\frac{\mathbf{1}\{t\in\mathcal{T}_{\mathcal{B}}\}}{|\mathcal{T}_{\mathcal{B}}|}
\right],
\end{equation}
where the ratio is zero when $\mathcal{T}_{\mathcal{B}}$ is empty. Uniform within-class pair draws in Equation~\ref{eq:spatial-pair} satisfy
\begin{equation}
\mathbb{E}_{\mathcal{B},\,\mathrm{pair\ draws}}
\!\left[\mathcal{L}_{\mathrm{pair}}(\mathcal{B};\tau)\right]
=\sum_t w_t\overline{L}_{\tau,t}.
\label{eq:minibatch-pair-weighting}
\end{equation}
For any target date weights $\pi_t$ with $w_t>0$ whenever $\pi_t>0$, direct and selective ranking consequently satisfy
\begin{equation}
\sum_t\pi_t\frac{q_t^B}{P_t}
\leq
\frac{\beta_{\mathrm{rank}}}{\log 2}
\left(\max_{t:\pi_t>0}\frac{\pi_tN_t}{w_t}\right)
\mathbb{E}_{\mathcal{B},\,\mathrm{pair\ draws}}
\!\left[\mathcal{L}_{\mathrm{pair}}(\mathcal{B};\tau)\right].
\label{eq:sampled-daily-budget-bound}
\end{equation}
\end{proposition}

Exchangeability makes every full-date positive--negative pair receive the same expected coefficient, while the probability that a date is eligible and the normalization by $|\mathcal{T}_{\mathcal{B}}|$ produce $w_t$. Equation~\ref{eq:sampled-daily-budget-bound} exposes the mismatch with uniformly weighted macro daily recall: the constant depends on both date eligibility and $N_t$ and can be large. It is a finite-sample statement for the training dates, not a population guarantee for an unseen year. Training-mode batch normalization also makes a score depend on the other examples in its realized batch; in that setting Equation~\ref{eq:minibatch-pair-weighting} characterizes the sampler's induced weights but does not identify SGD with optimization of one fixed full-pair empirical loss.

\subsection{Sensitivity to the pairwise target}
\label{sec:when-reference-helps}

All members share the same pair supervision and differ only through the target $\alpha\Delta_{\mathrm{ref}}$, so they separate only insofar as that target is appreciable on the scale set by $\beta_{\mathrm{rank}}$.

Proposition~\ref{prop:alpha-collapse} limits differences in scalar gap-gradient weights when $\beta_{\mathrm{rank}}|\Delta_{\mathrm{ref}}|$ is small. The selective attenuation bound additionally restricts its departure from RDPO to reference-wrong pairs. If there are no such pairs, its departure is zero; otherwise the truncated moment equals their probability times the conditional mean absolute gap. These statements concern loss anchoring, not the pretrained initialization shared by all variants. They do not establish equivalent parameter updates, stopping epochs, or ranking performance. Negative RDPO targets can be exceeded while the final pair remains misordered; all these logistic losses nevertheless continue to favor larger gaps.

Proposition~\ref{prop:alpha-collapse} controls the magnitude of scalar-weight differences, while Proposition~\ref{prop:ranking-gradient} describes their allocation across pair strata. Larger scaled pretrained margins allow larger differences between objectives. Task-level benefits are evaluated separately through daily AP and top-5\% recall; the bounds do not predict performance.

\subsection{Proof of Proposition~\ref{prop:alpha-collapse} and the measured margin scale}
\label{app:alpha-collapse}

From Proposition~\ref{prop:ranking-gradient}, $\partial g_\alpha/\partial\alpha=\beta_{\mathrm{rank}}\Delta_{\mathrm{ref}}g_\alpha(1-g_\alpha)$ for a detached reference. Since $g(1-g)\le1/4$ for $g\in(0,1)$, the mean value theorem gives Equation~\ref{eq:alpha-collapse}. Thus the scalar gradient factors lie in a band of width $\beta_{\mathrm{rank}}|\Delta_{\mathrm{ref}}|/4$ around any one member. This bound does not by itself constrain final ranking gains. At score-equal initialization $g_\alpha^{(0)}=\sigma[-\beta_{\mathrm{rank}}(1-\alpha)\Delta_{\mathrm{ref}}]$, so $g_0^{(0)}-g_1^{(0)}=\sigma(-\beta_{\mathrm{rank}}\Delta_{\mathrm{ref}})-1/2$; expanding $\sigma(-u)=1/2-u/4+u^3/48+O(u^5)$ gives the stated cubic remainder. For the selective target $g_{\mathrm{sel}}^{(0)}-g_1^{(0)}=[\sigma(-\beta_{\mathrm{rank}}\Delta_{\mathrm{ref}})-1/2]_+$, which vanishes whenever $\Delta_{\mathrm{ref}}\ge0$; bounding the surviving term by $\beta_{\mathrm{rank}}|\Delta_{\mathrm{ref}}|/4$ and taking expectations over pairs yields Equation~\ref{eq:selective-attenuation}. The reference-wrong event restricts selective's departure from RDPO; when its probability is zero, the indicator-moment bound is zero without an undefined conditional expectation.

The asymmetry between the two endpoints is also structural. For $\Delta_{\mathrm{ref}}\ll0$ the RDPO condition $\Delta_\theta>\Delta_{\mathrm{ref}}$ is satisfied at values with $\Delta_\theta<0$, so exceeding the RDPO target alone does not certify a corrected ordering; the logistic loss still favors further increases. Direct ranking retains the pretrained parameters and representation but omits the frozen pretrained model margin from its loss, including on pairs the reference already orders correctly. The selective target applies a zero target to reference-misranked pairs and retains positive reference margins elsewhere. These targets specify training emphasis; held-out pair accuracy and top-ranked event recall assess the resulting decisions.

\begin{table}[t]
\centering
\caption{Frozen pretrained score-gap scale, measured on held-out positive--negative pairs (cross-date for Yosemite, same-date on the fixed 20\% diagnostic sample for California) and averaged over seeds. $g_\alpha^{(0)}$ is the score-equal-initialization gradient weight of Equation~\ref{eq:ranking-initial-gradient}; RDPO fixes it at $1/2$ for every pair, and rel.\ expresses the direct--RDPO separation as a fraction of that $1/2$. Proposition~\ref{prop:alpha-collapse} bounds each mean separation by $\beta_{\rm rank}\mathbb{E}|\Delta_{\rm ref}|/4$. These are scalar gradient weights at a single model state: they bound how far the targets can differ, not whether any difference improves ranking. Table~\ref{tab:training-batch-margins} repeats the diagnostic on sampled-query diagnostic batches without assuming score equality.}
\label{tab:reference-margin-scale}
\begingroup
\fontsize{9}{11}\selectfont
\setlength{\tabcolsep}{4pt}
\begin{tabular}{@{}lrrrrrrr@{}}
\toprule
Setting & $\beta_{\rm rank}$ & $\mathbb{E}|\Delta_{\rm ref}|$ & $\beta_{\rm rank}\mathbb{E}|\Delta_{\rm ref}|$ & Ref.\ pair acc.\ (\%) & $\mathbb{E}|g_0^{(0)}{-}g_1^{(0)}|$ & rel.\ (\%) & $\mathbb{E}|g_{\rm sel}^{(0)}{-}g_1^{(0)}|$ \\
\midrule
Yosemite & 1 & 0.77 & 0.77 & 70.3 & 0.1568 & 31.4 & 0.0320 \\
California & 0.1 & 1.01 & 0.10 & 76.8 & 0.0253 & 5.1 & 0.0026 \\
\bottomrule
\end{tabular}
\endgroup
\end{table}

Table~\ref{tab:reference-margin-scale} evaluates these quantities on the frozen pretrained model using held-out positive--negative pairs: all $1{,}060{,}686$ cross-date pairs per seed for Yosemite, matching its cross-date training pairs, and all $355{,}227$ same-date pairs per seed from the fixed 20\% California diagnostic sample. This frozen-reference diagnostic is independent of the adaptation loss; all adapted-policy results use the complete grid. Yosemite logits are recovered from the stored probabilities. These are evaluation pairs, so they measure the reference's margin \emph{scale} rather than the exact training-pair distribution drawn from $\local$; no quantity here is fitted. California's measured $\mathbb{E}|g_0^{(0)}-g_1^{(0)}|=0.0253$ is close to the Equation~\ref{eq:alpha-collapse} bound $\beta_{\mathrm{rank}}\mathbb{E}|\Delta_{\mathrm{ref}}|/4=0.0253$ at the reported precision, consistent with a near-linear regime; at nonzero margins the Lipschitz bound is generally strict, so the displayed equality reflects rounding. Yosemite's $0.157$ falls further below its bound of $0.193$ as the sigmoid begins to saturate. Sweeping $\beta_{\mathrm{rank}}$ on the same references, the relative separation $\mathbb{E}|g_0^{(0)}-g_1^{(0)}|/(1/2)$ rises from $5.1\%$ at $\beta_{\mathrm{rank}}=0.1$ to $12.4\%$ at $0.25$, $23.2\%$ at $0.5$, $39.3\%$ at $1$ and $78.4\%$ at $5$. All percentages in this subsection are seed means, matching the table. A $\beta_{\mathrm{rank}}$ sweep would have to span roughly this range for the variants to differ appreciably, but changing $\beta_{\mathrm{rank}}$ rescales both losses rather than isolating the reference term, so direct ranking and RDPO require the same tuning opportunity for an interpretable comparison.

\paragraph{Real training batches without the score-equality assumption.}
Table~\ref{tab:reference-margin-scale} assumes policy and reference scores coincide at initialization. They do not: the policy runs training-mode and the frozen pretrained model evaluation-mode batch normalization, so $\Delta_{\theta,0}\neq\Delta_{\mathrm{ref}}$ even though the parameters are copied. Table~\ref{tab:training-batch-margins} therefore recomputes $g_{\mathrm{direct}}=\sigma(-\beta_{\mathrm{rank}}\Delta_{\theta,0})$, $g_{\mathrm{RDPO}}=\sigma[-\beta_{\mathrm{rank}}(\Delta_{\theta,0}-\Delta_{\mathrm{ref}})]$ and the selective analogue on California diagnostic batches from the original sampled-query retrieval set and pair draws at initialization, pooling $25$ batch draws ($3{,}751$ pairs per seed) with the parameters held at initialization. The realized departure is substantial, $\mathbb{E}|\Delta_{\theta,0}-\Delta_{\mathrm{ref}}|=0.28$, yet at the trained $\beta_{\mathrm{rank}}=0.1$ the realized mean separations, $0.0237$ for direct--RDPO and $0.0013$ for selective--RDPO, match their score-equal counterparts to the reported precision. The score-equal approximation thus matches the realized mean separations at the trained setting. The $99$th percentile of $|g_{\mathrm{direct}}-g_{\mathrm{RDPO}}|$ is $0.097$, showing larger differences on some pairs. Pretrained pair accuracy is $83.2\%$ on retrieved training pairs and $76.8\%$ on held-out pairs, which helps explain the smaller selective--RDPO separation in training. These measurements describe pairwise initialization-level scalar weights, with no focal gradient or parameter update; they do not report trained-policy performance for either retrieval setting.

\begin{table}[t]
\centering
\caption{Scalar pairwise gradient weights on California diagnostic batches from sampled-query retrieval at initialization; only pairwise gradients are measured. The adapted model uses training-mode batch normalization and the pretrained model evaluation mode. Measurements pool 25 batch draws (3,751 pairs per seed) and average model seeds 11, 22, 33; no parameters are updated. Although $\mathbb{E}|\Delta_{\theta,0}-\Delta_{\rm ref}|=0.28$, realized mean separations match the score-equal approximation at the reported precision. $\dagger$ marks the trained $\beta_{\rm rank}$. The sweep describes scalar-weight differences; changing $\beta_{\rm rank}$ reshapes both losses.}
\label{tab:training-batch-margins}
\begingroup
\fontsize{9}{11}\selectfont
\setlength{\tabcolsep}{4pt}
\begin{tabular}{@{}lrrrrrr@{}}
\toprule
& \multicolumn{2}{c}{$|g_{\rm direct}-g_{\rm RDPO}|$} & \multicolumn{2}{c}{$|g_{\rm sel}-g_{\rm RDPO}|$} & Score-equal & Bound \\
\cmidrule(lr){2-3}\cmidrule(lr){4-5}
$\beta_{\rm rank}$ & mean & p99 & mean & p99 & mean & $\beta_{\rm rank}\mathbb{E}|\Delta_{\rm ref}|/4$ \\
\midrule
0.1$^{\dagger}$ & 0.0237 & 0.0974 & 0.0013 & 0.0203 & 0.0237 & 0.0238 \\
0.25 & 0.0580 & 0.2297 & 0.0033 & 0.0503 & 0.0580 & 0.0594 \\
0.5 & 0.1095 & 0.3968 & 0.0065 & 0.0992 & 0.1091 & 0.1188 \\
1 & 0.1881 & 0.5982 & 0.0126 & 0.1880 & 0.1862 & 0.2377 \\
2 & 0.2829 & 0.8180 & 0.0226 & 0.3297 & 0.2782 & 0.4754 \\
5 & 0.3928 & 0.9846 & 0.0388 & 0.5872 & 0.3874 & 1.1884 \\
\bottomrule
\end{tabular}
\endgroup
\end{table}

\subsection{Pretrained score structure and scale}

The relevant pair distribution is the one induced by the candidate task: for daily prioritization, pairs should share a date. Reference accuracy can be summarized by $\Pr(\Delta_{\mathrm{ref}}>0)+\tfrac12\Pr(\Delta_{\mathrm{ref}}=0)$, supplemented by performance near the inspection cutoff and margin diagnostics stratified by correct, wrong, and tied pairs. Aggregate AUROC does not fix those conditional margin distributions. Replacing $r$ by $a r+b$ for $a>0$ leaves its ordering unchanged but replaces $\beta_{\mathrm{rank}}\Delta_{\mathrm{ref}}$ by $a\beta_{\mathrm{rank}}\Delta_{\mathrm{ref}}$. Matching AUROC therefore does not control the reference-dependent objective or its gradients. With score-equal initialization, changing the common initial policy/reference score scale also changes $g_\alpha^{(0)}$ for $\alpha<1$.

\section{Experimental configuration and metric verification}
\label{app:implementation}

\subsection{Yosemite data and architecture details}

The full Yosemite file has 70,491 rows from 1997--2022 and 9,219 positives. The nominal protocol starts in 2001, but no 2001 rows exist, so effective training begins in 2002. The twelve inputs are maximum/minimum temperature and relative humidity, precipitation, wind speed/direction, vapor-pressure deficit, energy-release component, burning index, and 100-hour/1000-hour fuel moisture. Labels satisfy the supplied smoothed $\mathrm{DM}>0$; coordinates, time, dry matter, and carbon are excluded from inputs. Standardization fits only historical training rows. Retrieval uses the complete unlabeled target batch.

The pretrained TabNet has widths 8, four steps, $\gamma=1.3$, sparsemax, two independent/shared gated linear units, and virtual batch size 128. It uses 50 epochs, Adam at $5\times10^{-3}$, and batches of 4096. Historical F1 selects each model's classification threshold.

\subsection{From-scratch joint focal and direct ranking}
\label{app:joint-training}

The joint baseline initializes a fresh TabNet with the same architecture and seed as the pretrained focal model and trains on the complete historical window. Its objective is
\begin{equation}
\mathcal{L}_{\mathrm{joint}}=\mathcal{L}_{\mathrm{focal}}+
\mathbb{E}_{(x^+,x^-)}\operatorname{softplus}(-\Delta_\theta)+0.05M_{\mathrm{loss}}.
\end{equation}
Focal supervision uses $\gamma=2$ and positive-class weight $1-\pi_+$, where $\pi_+$ is historical training prevalence. Direct ranking uses coefficient one and $\beta_{\mathrm{rank}}=1$, with every positive--negative pair in each shuffled minibatch, including cross-date pairs. We retain the pretraining Adam optimizer at $5\times10^{-3}$, batch size 4096, 50 epochs, and final checkpoint. The mask term remains zero. No reference scores, pretrained policy weights, or target-dependent retrieval enter joint optimization. Historical training F1 chooses its classification threshold; target labels enter only final evaluation.

This tests one global joint-training configuration against subsequent local adaptation. Training data, optimizer, initialization, and update counts differ from the local policies, so the comparison does not isolate initialization alone or match total computation. The from-scratch joint baseline uses the same training schedule as the pretrained focal model; the local methods additionally perform 100 adaptation epochs. The initial-gradient analysis assumes pretrained initialization and therefore does not describe the joint baseline's random starting point. The rolling 2021 results reuse the corresponding joint executions after verifying identical data windows and reference comparison checkpoints.

\subsection{Pairwise policy details}

All three local ranking configurations use $\beta=1$, preference coefficient one, and no focal supervision during adaptation. Direct and selective ranking use every positive--negative input pair within a shuffled minibatch, including cross-date pairs. Binary-label RDPO compares the true and complementary labels for each input, with no extra positive-label term ($\lambda_2=0$). Each retains the same $+0.05M_{\mathrm{loss}}$ term, whose realized value is zero. The configurations share pretrained checkpoints, retrieved rows, seeded minibatches, AdamW settings, and 100 adaptation epochs. Their candidate definitions differ, so this comparison does not isolate only the cross-input reference target. No interpolation sweep is evaluated.

The weighted-focal control retains fixed weights $c_0=2$ and $c_1=2+0.1/\pi_+$ on the aligned focal loss. It is a separate static reweighting baseline; it does not match the pure binary-label RDPO coefficients.

Gradient matching is an auxiliary scaling control. The actual TabNet mask term is zero in these experiments, and AdamW approximately cancels uniform loss scaling; near-identical fixed/matched results should therefore not be overinterpreted as a broad robustness result.

\subsection{Rolling evaluation and metric computation}

The rolling aggregate and daily-ranking tables report means $\pm$ 95\% Student-$t$ seed-CI half-widths. For rolling, average the three test years separately within each seed, then compute $t_{0.975,n-1}s/\sqrt{n}$ across $n=3$ seed averages; the 2021 daily table uses $n=5$ seed scores.

In 2021, 57 dates contain positives, of which 42 have both labels. With nine candidates per day, $\lceil0.05\times9\rceil=1$: the nominal daily Top5\% metric is Top1 at an effective selection fraction of 11.11\%. Mixed-day AP excludes all-positive dates. We report seed variability and paired directions without row-level significance tests.

\subsection{Fixed configurations}

\begin{table}[t]
\centering
\caption{Shared Yosemite configuration and objective-specific differences.}
\label{tab:configuration}
\begingroup
\papertablefont
\setlength{\tabcolsep}{4pt}
\renewcommand{\arraystretch}{1.12}
\begin{tabularx}{\linewidth}{@{}>{\raggedright\arraybackslash}p{130pt}X@{}}
\toprule
Component & Value \\
\midrule
Features / pretrained model & 12 weather variables / TabNet, $n_d=n_a=8$, 4 steps \\
Pretraining objective / optimizer & focal, $\gamma=2$ / Adam, lr $5\times10^{-3}$ \\
Pretraining epochs & 50 \\
Retrieval & training-standardized Euclidean KNN, $k=5$ \\
Local optimizer / epochs / batch & AdamW, lr $10^{-4}$ / 100 / 4096 \\
binary-label RDPO $\beta,\lambda_1,\lambda_2$ & $1.0,1.0,0.0$ \\
Pairwise ranking $\beta_{\rm rank}$ & 1.0 \\
Ranking adaptation focal term & absent for direct, selective, and binary-label RDPO \\
Sparsity term & $+0.05M_{\rm loss}$ (zero in these runs) \\
Main / rolling seeds & $\{11,17,23,29,37\}$ / $\{11,17,23\}$ \\
Checkpoint / threshold & final epoch / historical-training F1 optimum \\
Joint initialization / training set & random / all historical rows \\
Joint optimizer / epochs / coefficients & Adam, lr $5\times10^{-3}$ / 50 / focal 1, direct ranking 1 \\
Direct / fixed selective coefficient & 1.0 / 1.0 \\
Selective target & $[\Delta_{\rm ref}]_+$, detached, no cap \\
\bottomrule
\end{tabularx}
\endgroup
\end{table}

The 2021 local set has 11,187 unique historical rows and 1,582 positives ($\pi_+=0.1414$), identical across policies and seeds. The TabNet group-attention matrix contains one all-ones group, yielding a zero mask-loss term throughout training. The sparsity term enters the objective with coefficient $+0.05$.

\subsection{Pair identity and gradient matching}

The direct and selective runs use exactly the same retrieved historical examples and seeded minibatch orders. Every minibatch contributes the full Cartesian product of its positive and negative rows; empty-class batches would be skipped. Both methods have the same per-epoch pair counts and no skipped batches. Selective reference logits are cached in evaluation mode and detached. Policy training mode retains the original batch-normalization behavior.

For each seed, an independent copy reproduces the first actual training batch without advancing the training shuffle generator. Let $G_{\mathrm{direct}}$ and $G_{\mathrm{sel}}$ be the gradients of the unweighted pairwise terms with respect to all trainable parameters. The matched coefficient is
\begin{equation}
\lambda_{\mathrm{match}}=\frac{\|G_{\mathrm{direct}}\|_2}{\|G_{\mathrm{sel}}\|_2}.
\end{equation}
The two norms are finite and nonzero, and the direct coefficient is one. The initial weighted norm then agrees exactly, but later gradients and optimization trajectories need not agree. Table~\ref{tab:gradient-matching} reports measured norms. Because the other loss term is zero, this control largely scales the entire objective; AdamW's normalization makes its predictions almost identical to fixed-coefficient selective. The control therefore addresses initial scaling rather than broader objective sensitivity.

\begin{table}[t]
\centering
\caption{Initial ranking parameter-gradient norms on the first actual minibatch. Matched coefficient: $\|G_{\rm direct}\|_2/\|G_{\rm selective}\|_2$. Rolling 2021 reuses the identical main executions.}
\label{tab:gradient-matching}
\begingroup
\papertablefont
\setlength{\tabcolsep}{2pt}
\renewcommand{\arraystretch}{1.12}
\renewcommand{\tabularxcolumn}[1]{m{#1}}
\begin{tabularx}{\linewidth}{@{}>{\centering\arraybackslash}X>{\centering\arraybackslash}X>{\centering\arraybackslash}X>{\centering\arraybackslash}X>{\centering\arraybackslash}X>{\centering\arraybackslash}X@{}}
\toprule
\textbf{Year} & \textbf{Seed} & \textbf{$\|G_{\rm direct}\|_2$} & \textbf{$\|G_{\rm selective}\|_2$} & \textbf{Coefficient} & \textbf{Pairs} \\
\midrule
2021 & 11 & 0.1702 & 0.5082 & 0.3350 & 2000943 \\\addlinespace[2.5pt]
2021 & 17 & 0.1896 & 0.4687 & 0.4045 & 2155120 \\\addlinespace[2.5pt]
2021 & 23 & 0.1409 & 0.4714 & 0.2989 & 2012775 \\\addlinespace[2.5pt]
2021 & 29 & 0.2067 & 0.5316 & 0.3888 & 1938300 \\\addlinespace[2.5pt]
2021 & 37 & 0.1939 & 0.6093 & 0.3181 & 2045148 \\
\midrule
2019 & 11 & 0.2927 & 1.3228 & 0.2213 & 1688415 \\\addlinespace[2.5pt]
2019 & 17 & 0.2834 & 1.1193 & 0.2532 & 1763823 \\\addlinespace[2.5pt]
2019 & 23 & 0.2955 & 1.1031 & 0.2679 & 1865628 \\
\midrule
2020 & 11 & 0.1364 & 0.4049 & 0.3369 & 1997980 \\\addlinespace[2.5pt]
2020 & 17 & 0.1273 & 0.3217 & 0.3957 & 2051008 \\\addlinespace[2.5pt]
2020 & 23 & 0.1383 & 0.3267 & 0.4235 & 1868679 \\
\bottomrule
\end{tabularx}
\endgroup
\end{table}

\subsection{Evaluation and reuse}

Each classification threshold maximizes F1 over all historical training predictions. For global recall at fraction $q$, the model selects $\lceil qm\rceil$ target grid--days and divides captured positives by all target positives. Daily diagnostics instead select $\max\{1,\lceil0.05|\mathcal{C}_t|\rceil\}$ cells on each date; stable score ties follow original candidate-row order, without labels. Daily AP uses exact score-tie handling and excludes zero-positive dates; mixed-day AP further excludes all-positive dates. Reference-state diagnostics use strict positive/negative reference gaps; tied pairs are counted separately in the diagnostics.

The reused rolling 2021 cases have identical frozen pretrained models, adapted model states, retrieved examples, data partitions, and hyperparameters to their corresponding main executions. The five-seed and rolling summaries therefore contain overlapping 2021 cases. Classification, daily-ranking, and pair-stratum metrics are independently recomputed from final predictions. All training losses and prediction scores are finite; no margin cap is used.

\section{Additional Yosemite results}
\label{app:results}

The Yosemite result tables use a fixed 9pt font. Bold marks the best displayed mean in each column, including ties, within each experiment group (and pair scope where applicable). Larger is better except for Miss, where smaller is better.

\subsection{Daily risk-ranking diagnostics}

Table~\ref{tab:daily-ranking} reports daily candidate-ordering metrics with 95\% Student-$t$ confidence intervals over seeds. For 2021 there are five seeds; rolling intervals use the three per-seed averages across the fixed 2019--2021 test years. These intervals describe training randomness conditional on the evaluated data, rather than uncertainty over future years. Daily AP and Top1 recall complement pooled ranking by assessing the within-date candidate task.

\begin{table}[t]
\centering
\caption{Daily ranking: mean $\pm$ 95\% Student-$t$ seed-CI half-width. Use five seeds for 2021; for rolling, average three years within each of three seeds. CIs condition on these test years. AP excludes zero-fire days; mixed AP also excludes all-positive days. Top1 is one of nine cells (11.11\% budget). All values are percentages.}
\label{tab:daily-ranking}
\begingroup
\papertablefont
\setlength{\tabcolsep}{2pt}
\renewcommand{\arraystretch}{1.12}
\renewcommand{\tabularxcolumn}[1]{m{#1}}
\begin{tabularx}{\linewidth}{@{}>{\raggedright\arraybackslash}m{53pt}>{\raggedright\arraybackslash}m{70pt}>{\centering\arraybackslash}X>{\centering\arraybackslash}X>{\centering\arraybackslash}X>{\centering\arraybackslash}X@{}}
\toprule
\textbf{Experiment} & \textbf{Method} & \textbf{Daily AP} & \textbf{Mixed AP} & \textbf{R@Top1} & \textbf{NDCG@Top1} \\
\midrule
2021 & Pretrained & 77.24\,$\pm$\,1.54 & 69.11\,$\pm$\,2.09 & 9.24\,$\pm$\,1.24 & 59.65\,$\pm$\,6.35 \\\addlinespace[2.5pt]
 & Joint focal + direct & 77.40\,$\pm$\,0.71 & 69.33\,$\pm$\,0.97 & 9.04\,$\pm$\,0.68 & 58.95\,$\pm$\,3.96 \\\addlinespace[2.5pt]
 & Direct ranking & 77.38\,$\pm$\,1.37 & 69.30\,$\pm$\,1.86 & 9.24\,$\pm$\,1.09 & 60.00\,$\pm$\,6.43 \\\addlinespace[2.5pt]
 & binary-label RDPO & \textbf{78.12}\,$\pm$\,0.27 & \textbf{70.31}\,$\pm$\,0.37 & \textbf{10.03}\,$\pm$\,0.89 & \textbf{64.21}\,$\pm$\,5.68 \\\addlinespace[2.5pt]
 & Selective ranking & 77.78\,$\pm$\,0.93 & 69.85\,$\pm$\,1.26 & 9.39\,$\pm$\,0.91 & 60.70\,$\pm$\,5.47 \\
\midrule
Rolling & Pretrained & 81.68\,$\pm$\,3.58 & 73.10\,$\pm$\,4.55 & 11.68\,$\pm$\,2.19 & 72.20\,$\pm$\,9.73 \\\addlinespace[2.5pt]
 & Joint focal + direct & 80.80\,$\pm$\,3.44 & 72.23\,$\pm$\,2.97 & 11.50\,$\pm$\,1.07 & 71.37\,$\pm$\,5.87 \\\addlinespace[2.5pt]
 & Direct ranking & \textbf{82.56}\,$\pm$\,3.49 & 72.96\,$\pm$\,3.69 & 11.95\,$\pm$\,0.58 & 74.09\,$\pm$\,2.35 \\\addlinespace[2.5pt]
 & binary-label RDPO & 81.97\,$\pm$\,5.69 & \textbf{73.41}\,$\pm$\,5.91 & \textbf{11.99}\,$\pm$\,1.83 & \textbf{74.52}\,$\pm$\,8.90 \\\addlinespace[2.5pt]
 & Selective ranking & 82.52\,$\pm$\,5.59 & 72.97\,$\pm$\,5.30 & 11.68\,$\pm$\,1.10 & 72.97\,$\pm$\,5.59 \\
\bottomrule
\end{tabularx}
\endgroup
\end{table}

\subsection{Complete classification metrics}

Tables~\ref{tab:complete-2021} and \ref{tab:classification-rolling} supplement the global ranking tables with accuracy, recall, F1, and miss rates. Each model uses its own historically selected F1 operating point. These threshold metrics complement the ranking comparisons at the same selected fraction.

\begin{table}[t]
\centering
\caption{2021 classification results, five seeds. Mean $\pm$ SD; percentages. Miss is one minus recall.}
\label{tab:complete-2021}
\begingroup
\papertablefont
\setlength{\tabcolsep}{2pt}
\renewcommand{\arraystretch}{1.12}
\renewcommand{\tabularxcolumn}[1]{m{#1}}
\begin{tabularx}{\linewidth}{@{}>{\raggedright\arraybackslash}m{70pt}>{\centering\arraybackslash}X>{\centering\arraybackslash}X>{\centering\arraybackslash}X>{\centering\arraybackslash}X@{}}
\toprule
\textbf{Method} & \textbf{Accuracy} & \textbf{Recall} & \textbf{F1} & \textbf{Miss} \\
\midrule
Pretrained & 71.09\,$\pm$\,2.15 & 57.80\,$\pm$\,5.70 & 30.61\,$\pm$\,0.75 & 42.20\,$\pm$\,5.70 \\\addlinespace[2.5pt]
Joint focal + direct & \textbf{72.12}\,$\pm$\,1.75 & 52.73\,$\pm$\,6.77 & 29.39\,$\pm$\,1.77 & 47.27\,$\pm$\,6.77 \\\addlinespace[2.5pt]
Continued focal & 70.59\,$\pm$\,1.63 & 57.02\,$\pm$\,6.80 & 29.93\,$\pm$\,1.91 & 42.98\,$\pm$\,6.80 \\\addlinespace[2.5pt]
Weighted focal & 71.09\,$\pm$\,1.35 & 56.09\,$\pm$\,6.07 & 29.94\,$\pm$\,1.56 & 43.91\,$\pm$\,6.07 \\\addlinespace[2.5pt]
Direct ranking & 70.67\,$\pm$\,0.83 & 60.50\,$\pm$\,2.37 & \textbf{31.31}\,$\pm$\,1.11 & 39.50\,$\pm$\,2.37 \\\addlinespace[2.5pt]
binary-label RDPO & 69.60\,$\pm$\,2.15 & 61.27\,$\pm$\,5.86 & 30.79\,$\pm$\,1.41 & 38.73\,$\pm$\,5.86 \\\addlinespace[2.5pt]
Selective ranking & 70.19\,$\pm$\,1.40 & \textbf{61.32}\,$\pm$\,3.50 & 31.25\,$\pm$\,0.86 & \textbf{38.68}\,$\pm$\,3.50 \\
\bottomrule
\end{tabularx}
\endgroup
\end{table}

\begin{table}[t]
\centering
\caption{Rolling classification results. Means over nine year--seed cases; percentages.}
\label{tab:classification-rolling}
\begingroup
\papertablefont
\setlength{\tabcolsep}{2pt}
\renewcommand{\arraystretch}{1.12}
\renewcommand{\tabularxcolumn}[1]{m{#1}}
\begin{tabularx}{\linewidth}{@{}>{\raggedright\arraybackslash}m{70pt}>{\centering\arraybackslash}X>{\centering\arraybackslash}X>{\centering\arraybackslash}X>{\centering\arraybackslash}X@{}}
\toprule
\textbf{Method} & \textbf{Accuracy} & \textbf{Recall} & \textbf{F1} & \textbf{Miss} \\
\midrule
Pretrained & 74.63 & 56.01 & 29.29 & 43.99 \\\addlinespace[2.5pt]
Joint focal + direct & \textbf{75.36} & 50.97 & 28.16 & 49.03 \\\addlinespace[2.5pt]
Continued focal & 73.50 & 59.05 & 29.36 & 40.95 \\\addlinespace[2.5pt]
Weighted focal & 73.63 & 58.57 & 29.35 & 41.43 \\\addlinespace[2.5pt]
Direct ranking & 72.52 & 64.61 & \textbf{30.60} & 35.39 \\\addlinespace[2.5pt]
binary-label RDPO & 70.83 & 60.24 & 28.17 & 39.76 \\\addlinespace[2.5pt]
Selective ranking & 72.19 & \textbf{65.29} & 30.55 & \textbf{34.71} \\
\bottomrule
\end{tabularx}
\endgroup
\end{table}

\subsection{Complete rolling summaries}

Table~\ref{tab:rolling-recall} reports annual recall for the seven policies as mean $\pm$ SD over three seeds per year. Table~\ref{tab:rolling-overall} preserves the numerical comparisons shown in Figure~\ref{fig:rolling-ranking}, including 95\% seed confidence intervals for all six metrics.

\begin{table}[t]
\centering
\caption{Recall by test year, three seeds per year. Mean $\pm$ SD; percentages. The 2021 column uses three of the five main seeds.}
\label{tab:rolling-recall}
\begingroup
\papertablefont
\setlength{\tabcolsep}{2pt}
\renewcommand{\arraystretch}{1.12}
\renewcommand{\tabularxcolumn}[1]{m{#1}}
\begin{tabularx}{\linewidth}{@{}>{\raggedright\arraybackslash}m{70pt}>{\centering\arraybackslash}X>{\centering\arraybackslash}X>{\centering\arraybackslash}X@{}}
\toprule
\textbf{Method} & \textbf{2019} & \textbf{2020} & \textbf{2021} \\
\midrule
Pretrained & 38.41\,$\pm$\,9.37 & 74.07\,$\pm$\,9.65 & 55.56\,$\pm$\,6.75 \\\addlinespace[2.5pt]
Joint focal + direct & 37.32\,$\pm$\,7.86 & 61.79\,$\pm$\,9.84 & 53.81\,$\pm$\,5.82 \\\addlinespace[2.5pt]
Continued focal & 46.01\,$\pm$\,4.39 & 76.23\,$\pm$\,9.24 & 54.91\,$\pm$\,7.99 \\\addlinespace[2.5pt]
Weighted focal & 45.29\,$\pm$\,5.14 & 76.23\,$\pm$\,9.56 & 54.18\,$\pm$\,7.50 \\\addlinespace[2.5pt]
Direct ranking & 51.09\,$\pm$\,9.48 & 82.96\,$\pm$\,2.77 & 59.78\,$\pm$\,2.90 \\\addlinespace[2.5pt]
binary-label RDPO & 48.91\,$\pm$\,8.90 & 74.51\,$\pm$\,4.39 & 57.30\,$\pm$\,2.98 \\\addlinespace[2.5pt]
Selective ranking & \textbf{51.81}\,$\pm$\,12.69 & \textbf{83.83}\,$\pm$\,3.80 & \textbf{60.24}\,$\pm$\,4.43 \\
\bottomrule
\end{tabularx}
\endgroup
\end{table}

\begin{table}[t]
\centering
\caption{Rolling evaluation, nine year--seed cases. Mean $\pm$ 95\% seed-CI half-width, conditional on these test years: average the three years within each seed, then use a Student-$t$ interval over three seeds. Percentages; bold marks the largest displayed mean.}
\label{tab:rolling-overall}
\begingroup
\papertablefont
\setlength{\tabcolsep}{2pt}
\renewcommand{\arraystretch}{1.12}
\renewcommand{\tabularxcolumn}[1]{m{#1}}
\begin{tabularx}{\linewidth}{@{}>{\raggedright\arraybackslash}m{70pt}>{\centering\arraybackslash}X>{\centering\arraybackslash}X>{\centering\arraybackslash}X>{\centering\arraybackslash}X>{\centering\arraybackslash}X>{\centering\arraybackslash}X@{}}
\toprule
\textbf{Method} & \textbf{Recall} & \textbf{F1} & \textbf{AUROC} & \textbf{AUPRC} & \textbf{R@20} & \textbf{R@30} \\
\midrule
Pretrained & 56.01\,$\pm$\,19.65 & 29.29\,$\pm$\,4.71 & 73.12\,$\pm$\,4.16 & 22.85\,$\pm$\,6.82 & 43.03\,$\pm$\,9.38 & 60.81\,$\pm$\,13.20 \\\addlinespace[2.5pt]
Joint focal + direct & 50.97\,$\pm$\,9.48 & 28.16\,$\pm$\,1.70 & 71.73\,$\pm$\,0.97 & 21.43\,$\pm$\,2.19 & 42.00\,$\pm$\,5.91 & 57.06\,$\pm$\,1.82 \\\addlinespace[2.5pt]
Continued focal & 59.05\,$\pm$\,16.11 & 29.36\,$\pm$\,3.96 & 73.08\,$\pm$\,3.97 & 22.08\,$\pm$\,4.57 & 43.09\,$\pm$\,7.28 & 60.32\,$\pm$\,11.63 \\\addlinespace[2.5pt]
Weighted focal & 58.57\,$\pm$\,16.40 & 29.35\,$\pm$\,3.31 & 73.06\,$\pm$\,3.92 & 22.00\,$\pm$\,4.55 & 42.93\,$\pm$\,7.76 & 60.11\,$\pm$\,11.45 \\\addlinespace[2.5pt]
Direct ranking & 64.61\,$\pm$\,12.17 & \textbf{30.60}\,$\pm$\,2.32 & 74.19\,$\pm$\,2.21 & 24.30\,$\pm$\,2.60 & 45.41\,$\pm$\,0.46 & 62.96\,$\pm$\,6.63 \\\addlinespace[2.5pt]
binary-label RDPO & 60.24\,$\pm$\,12.70 & 28.17\,$\pm$\,2.30 & 72.13\,$\pm$\,3.19 & 22.05\,$\pm$\,6.00 & 41.72\,$\pm$\,9.17 & 57.00\,$\pm$\,8.07 \\\addlinespace[2.5pt]
Selective ranking & \textbf{65.29}\,$\pm$\,16.92 & 30.55\,$\pm$\,2.47 & \textbf{74.32}\,$\pm$\,1.96 & \textbf{24.35}\,$\pm$\,2.89 & \textbf{45.79}\,$\pm$\,4.08 & \textbf{63.26}\,$\pm$\,6.53 \\
\bottomrule
\end{tabularx}
\endgroup
\end{table}

\subsection{Year-specific ranking and paired differences}

Direct and selective ranking have similar rolling pooled means: AUROC of 74.19\% and 74.32\%, AUPRC of 24.30\% and 24.35\%, and daily AP of 82.56\% and 82.52\%, respectively. Tables~\ref{tab:rolling-ranking}--\ref{tab:rolling-r30} give annual AUROC, AUPRC, R@20, and R@30. Each cell reports the three-seed mean $\pm$ SD. Target sets contain 92 positives in 2019, 540 in 2020, and 363 in 2021 among approximately 3,285 grid--days. Changing prevalence strongly affects AUPRC, so compare methods within a year. The overall rolling means weight each year--seed case equally. Selective improves AUROC over direct ranking in 7/9 cases and AUPRC in 6/9; the tables retain all annual comparisons.

\begin{table}[t]
\centering
\caption{AUROC by rolling test year, three seeds per year. Mean $\pm$ SD; percentages.}
\label{tab:rolling-ranking}
\begingroup
\papertablefont
\setlength{\tabcolsep}{2pt}
\renewcommand{\arraystretch}{1.12}
\renewcommand{\tabularxcolumn}[1]{m{#1}}
\begin{tabularx}{\linewidth}{@{}>{\raggedright\arraybackslash}m{70pt}>{\centering\arraybackslash}X>{\centering\arraybackslash}X>{\centering\arraybackslash}X@{}}
\toprule
\textbf{Method} & \textbf{2019} & \textbf{2020} & \textbf{2021} \\
\midrule
Pretrained & 67.91\,$\pm$\,2.00 & 81.33\,$\pm$\,2.71 & 70.12\,$\pm$\,0.60 \\\addlinespace[2.5pt]
Joint focal + direct & 66.95\,$\pm$\,2.12 & 78.88\,$\pm$\,2.64 & 69.38\,$\pm$\,1.41 \\\addlinespace[2.5pt]
Continued focal & 68.32\,$\pm$\,2.04 & 81.23\,$\pm$\,1.58 & 69.69\,$\pm$\,1.22 \\\addlinespace[2.5pt]
Weighted focal & 68.37\,$\pm$\,2.06 & 81.27\,$\pm$\,1.62 & 69.54\,$\pm$\,1.18 \\\addlinespace[2.5pt]
Direct ranking & 69.09\,$\pm$\,1.87 & \textbf{82.63}\,$\pm$\,0.62 & 70.85\,$\pm$\,0.46 \\\addlinespace[2.5pt]
binary-label RDPO & 66.95\,$\pm$\,1.73 & 78.90\,$\pm$\,1.89 & 70.53\,$\pm$\,0.41 \\\addlinespace[2.5pt]
Selective ranking & \textbf{69.13}\,$\pm$\,1.86 & 82.60\,$\pm$\,0.77 & \textbf{71.23}\,$\pm$\,0.32 \\
\bottomrule
\end{tabularx}
\endgroup
\end{table}

\begin{table}[t]
\centering
\caption{AUPRC by rolling test year, three seeds per year. Mean $\pm$ SD; percentages.}
\label{tab:rolling-auprc}
\begingroup
\papertablefont
\setlength{\tabcolsep}{2pt}
\renewcommand{\arraystretch}{1.12}
\renewcommand{\tabularxcolumn}[1]{m{#1}}
\begin{tabularx}{\linewidth}{@{}>{\raggedright\arraybackslash}m{70pt}>{\centering\arraybackslash}X>{\centering\arraybackslash}X>{\centering\arraybackslash}X@{}}
\toprule
\textbf{Method} & \textbf{2019} & \textbf{2020} & \textbf{2021} \\
\midrule
Pretrained & 4.35\,$\pm$\,0.30 & 43.69\,$\pm$\,7.45 & 20.50\,$\pm$\,0.65 \\\addlinespace[2.5pt]
Joint focal + direct & 4.34\,$\pm$\,0.50 & 39.98\,$\pm$\,3.54 & 19.98\,$\pm$\,0.92 \\\addlinespace[2.5pt]
Continued focal & 4.43\,$\pm$\,0.31 & 41.71\,$\pm$\,4.07 & 20.09\,$\pm$\,1.18 \\\addlinespace[2.5pt]
Weighted focal & 4.44\,$\pm$\,0.32 & 41.50\,$\pm$\,4.19 & 20.05\,$\pm$\,1.09 \\\addlinespace[2.5pt]
Direct ranking & 4.52\,$\pm$\,0.42 & \textbf{46.08}\,$\pm$\,3.07 & 22.29\,$\pm$\,0.31 \\\addlinespace[2.5pt]
binary-label RDPO & 4.15\,$\pm$\,0.24 & 40.53\,$\pm$\,6.87 & 21.47\,$\pm$\,0.49 \\\addlinespace[2.5pt]
Selective ranking & \textbf{4.56}\,$\pm$\,0.47 & 45.90\,$\pm$\,3.78 & \textbf{22.57}\,$\pm$\,0.60 \\
\bottomrule
\end{tabularx}
\endgroup
\end{table}

\begin{table}[t]
\centering
\caption{R@20 by rolling test year, three seeds per year. Mean $\pm$ SD; percentages.}
\label{tab:rolling-r20}
\begingroup
\papertablefont
\setlength{\tabcolsep}{2pt}
\renewcommand{\arraystretch}{1.12}
\renewcommand{\tabularxcolumn}[1]{m{#1}}
\begin{tabularx}{\linewidth}{@{}>{\raggedright\arraybackslash}m{70pt}>{\centering\arraybackslash}X>{\centering\arraybackslash}X>{\centering\arraybackslash}X@{}}
\toprule
\textbf{Method} & \textbf{2019} & \textbf{2020} & \textbf{2021} \\
\midrule
Pretrained & 34.42\,$\pm$\,5.47 & 52.90\,$\pm$\,7.44 & 41.78\,$\pm$\,1.24 \\\addlinespace[2.5pt]
Joint focal + direct & 35.14\,$\pm$\,8.79 & 50.19\,$\pm$\,5.30 & 40.68\,$\pm$\,3.13 \\\addlinespace[2.5pt]
Continued focal & 35.87\,$\pm$\,3.92 & 53.09\,$\pm$\,2.65 & 40.31\,$\pm$\,2.56 \\\addlinespace[2.5pt]
Weighted focal & 36.23\,$\pm$\,5.47 & 53.09\,$\pm$\,3.24 & 39.49\,$\pm$\,2.03 \\\addlinespace[2.5pt]
Direct ranking & \textbf{36.96}\,$\pm$\,1.09 & 55.37\,$\pm$\,0.32 & 43.89\,$\pm$\,0.89 \\\addlinespace[2.5pt]
binary-label RDPO & 33.70\,$\pm$\,3.92 & 48.40\,$\pm$\,5.27 & 43.07\,$\pm$\,2.14 \\\addlinespace[2.5pt]
Selective ranking & 36.59\,$\pm$\,3.82 & \textbf{55.86}\,$\pm$\,1.54 & \textbf{44.90}\,$\pm$\,1.46 \\
\bottomrule
\end{tabularx}
\endgroup
\end{table}

\begin{table}[t]
\centering
\caption{R@30 by rolling test year, three seeds per year. Mean $\pm$ SD; percentages.}
\label{tab:rolling-r30}
\begingroup
\papertablefont
\setlength{\tabcolsep}{2pt}
\renewcommand{\arraystretch}{1.12}
\renewcommand{\tabularxcolumn}[1]{m{#1}}
\begin{tabularx}{\linewidth}{@{}>{\raggedright\arraybackslash}m{70pt}>{\centering\arraybackslash}X>{\centering\arraybackslash}X>{\centering\arraybackslash}X@{}}
\toprule
\textbf{Method} & \textbf{2019} & \textbf{2020} & \textbf{2021} \\
\midrule
Pretrained & 55.43\,$\pm$\,7.61 & 70.99\,$\pm$\,6.88 & 56.01\,$\pm$\,2.92 \\\addlinespace[2.5pt]
Joint focal + direct & 50.36\,$\pm$\,4.90 & 65.80\,$\pm$\,5.87 & 55.00\,$\pm$\,3.05 \\\addlinespace[2.5pt]
Continued focal & 55.07\,$\pm$\,6.64 & 71.17\,$\pm$\,3.85 & 54.73\,$\pm$\,3.98 \\\addlinespace[2.5pt]
Weighted focal & 54.71\,$\pm$\,6.99 & 71.36\,$\pm$\,3.44 & 54.27\,$\pm$\,3.86 \\\addlinespace[2.5pt]
Direct ranking & 56.16\,$\pm$\,6.99 & 75.43\,$\pm$\,1.13 & 57.30\,$\pm$\,1.46 \\\addlinespace[2.5pt]
binary-label RDPO & 49.64\,$\pm$\,3.14 & 65.25\,$\pm$\,5.21 & 56.11\,$\pm$\,1.57 \\\addlinespace[2.5pt]
Selective ranking & \textbf{56.52}\,$\pm$\,5.75 & \textbf{75.49}\,$\pm$\,2.33 & \textbf{57.76}\,$\pm$\,1.88 \\
\bottomrule
\end{tabularx}
\endgroup
\end{table}

\begin{table}[t]
\centering
\caption{Paired selective-minus-direct differences (percentage points), calculated before rounding.}
\label{tab:selective-deltas}
\begingroup
\papertablefont
\setlength{\tabcolsep}{2pt}
\renewcommand{\arraystretch}{1.12}
\renewcommand{\tabularxcolumn}[1]{m{#1}}
\begin{tabularx}{\linewidth}{@{}>{\raggedright\arraybackslash}m{53pt}>{\raggedright\arraybackslash}m{70pt}>{\centering\arraybackslash}X>{\centering\arraybackslash}X>{\centering\arraybackslash}X>{\centering\arraybackslash}X>{\centering\arraybackslash}X>{\centering\arraybackslash}X@{}}
\toprule
\textbf{Experiment} & \textbf{Variant} & \textbf{\shortstack{$\Delta$\\AUC}} & \textbf{\shortstack{$\Delta$\\AP}} & \textbf{\shortstack{$\Delta$\\R@20}} & \textbf{\shortstack{$\Delta$\\R@30}} & \textbf{\shortstack{$\Delta$\\Recall}} & \textbf{\shortstack{$\Delta$\\F1}} \\
\midrule
2021 & Selective ranking & \textbf{+0.27} & \textbf{+0.13} & \textbf{+0.66} & \textbf{+0.00} & \textbf{+0.83} & \textbf{-0.06} \\
\midrule
Rolling & Selective ranking & \textbf{+0.13} & \textbf{+0.05} & \textbf{+0.38} & \textbf{+0.29} & \textbf{+0.68} & \textbf{-0.06} \\
\bottomrule
\end{tabularx}
\endgroup
\end{table}

\subsection{Reference-correct and reference-wrong pairs}

Table~\ref{tab:pair-strata} conditions on reference gap sign, with equal-weight averages across cases. Pooled pairs match the cross-date training interpretation; same-day pairs assess the daily candidate task. More wrong-pair corrections can coexist with lower correct-pair retention. Margin expansion does not ensure crossing zero on a wrong pair, nor does aggregate correction ensure improved daily top-one recall.

\begin{table}[t]
\centering
\caption{Held-out pair diagnostics, averaged over five seeds (2021) or nine year--seed cases (rolling). Corrected/retained means policy gap $>0$; expanded means policy gap exceeds the reference gap on correct pairs. Reference ties are excluded. Percentages.}
\label{tab:pair-strata}
\begingroup
\papertablefont
\setlength{\tabcolsep}{2pt}
\renewcommand{\arraystretch}{1.12}
\renewcommand{\tabularxcolumn}[1]{m{#1}}
\begin{tabularx}{\linewidth}{@{}>{\raggedright\arraybackslash}m{53pt}>{\raggedright\arraybackslash}m{39pt}>{\raggedright\arraybackslash}m{70pt}>{\centering\arraybackslash}X>{\centering\arraybackslash}X>{\centering\arraybackslash}X@{}}
\toprule
\textbf{Experiment} & \textbf{Scope} & \textbf{Method} & \textbf{\shortstack{Wrong\\corrected}} & \textbf{\shortstack{Correct\\retained}} & \textbf{\shortstack{Correct\\expanded}} \\
\midrule
2021 & Pooled & Direct ranking & 12.02 & 96.23 & 91.80 \\\addlinespace[2.5pt]
 &  & Selective ranking & \textbf{14.45} & 95.59 & \textbf{92.28} \\
\midrule
2021 & Same day & Direct ranking & 8.87 & 90.77 & \textbf{81.42} \\\addlinespace[2.5pt]
 &  & Selective ranking & \textbf{10.59} & 89.78 & 81.39 \\
\midrule
Rolling & Pooled & Direct ranking & 16.71 & 95.67 & 90.34 \\\addlinespace[2.5pt]
 &  & Selective ranking & \textbf{18.54} & 95.13 & \textbf{91.14} \\
\midrule
Rolling & Same day & Direct ranking & 18.65 & 88.38 & 72.89 \\\addlinespace[2.5pt]
 &  & Selective ranking & \textbf{22.04} & 87.07 & \textbf{74.49} \\
\bottomrule
\end{tabularx}
\endgroup
\end{table}

\section{California data and experimental protocol}
\label{app:california}

\subsection{Spatially smoothed occurrence and fixed subsampling}

The pilot uses 666 GFED cells whose centers lie within California. Previous-day GridMET weather predicts the next day's label. Let $D_{t,j}$ denote the raw daily dry-matter estimate and $d_{ij}$ Euclidean distance between cell centers in latitude/longitude degrees. Following the supplied Yosemite smoothing convention,
\begin{equation}
W_{ij}=\frac{\exp[-d_{ij}^2/(2\sigma_g^2)]\mathbf{1}\{d_{ij}\le c_g\}}
{\sum_{j'}\exp[-d_{ij'}^2/(2\sigma_g^2)]\mathbf{1}\{d_{ij'}\le c_g\}},
\qquad Y_{t,i}=\mathbf{1}\!\left\{\sum_j W_{ij}D_{t,j}>0\right\},
\label{eq:ca-label-smoothing}
\end{equation}
with $\sigma_g=0.25^{\circ}$ and $c_g=0.4^{\circ}$. The support and normalization are restricted to the California grid. For nonnegative dry matter, the binary rule is exactly $Y_{t,i}=\mathbf{1}\{\exists j:d_{ij}\le c_g,\ D_{t,j}>0\}$ in real arithmetic. Thus the cutoff determines binary support; the Gaussian bandwidth changes smoothed amounts but not this zero-threshold occurrence support. A positive label can arise from a neighboring positive cell within the cutoff; this target represents neighborhood-smoothed occurrence, not an independently measured ignition or an extreme-fire category. On the full 2022 grid, smoothing increases prevalence from 0.902\% to 6.072\%. Distances in degrees do not define a constant physical length across a continental area; a broader-area implementation should reassess the smoothing scale in an appropriate geographic metric before comparing labels across latitudes.

The final training period is 2002--2021 and testing is 2022. Uniform sampling without replacement within each historical year--label stratum retains 20\% of training rows. This fixed seed-11 sample contains 972,892 rows, including 76,785 positives. Retrieval queries all 243,090 unlabeled target grid--days and evaluation covers all 666 cells on each of 365 dates. The 2022 evaluation contains 14,760 positives across 351 positive dates. Target outcomes are read for evaluation after the predictor-only candidate covariates are constructed; no target labels determine candidate membership or supervise adaptation.

\subsection{Causal cell-history features}

For prediction year $Y$, let $\mathcal{I}_{i,A}$ be the full available grid--days for cell $i$ whose years belong to a past-year set $A$. Define $S_{i,A}=\sum_{t\in\mathcal{I}_{i,A}}Y_{t,i}$ and $N_{i,A}=|\mathcal{I}_{i,A}|$. With $A_Y=\{2002,\ldots,Y-1\}$ (empty for $Y=2002$) and $\kappa_A>0$, the statewide past prevalence and shrunk rate are
\begin{equation}
\overline\pi_{<Y}=
\begin{cases}
\displaystyle\frac{\sum_i S_{i,A_Y}}{\sum_i N_{i,A_Y}},&\sum_i N_{i,A_Y}>0,\\
0.01,&\sum_i N_{i,A_Y}=0,
\end{cases}
\qquad
\widehat\pi_{i,A}=\frac{S_{i,A}+\kappa_A\overline\pi_{<Y}}{N_{i,A}+\kappa_A}.
\label{eq:ca-causal-rate}
\end{equation}
The all-history feature uses $A=\{2002,\ldots,Y-1\}$ and the three-year feature uses its last three years, both with $\kappa_A=365$ pseudo-days. For calendar month $m$, the monthly feature restricts the all-history counts to that month and uses 30 pseudo-days toward the same statewide all-month past prevalence. In 2002 the cold-start prior is 0.01. We also supply the history length $Y-2002$ and seasonality $[\sin(2\pi\operatorname{DOY}(t)/365.25),\cos(2\pi\operatorname{DOY}(t)/365.25)]$.

These counts are constructed from full, spatially smoothed historical labels before subsampling, using only years strictly before each input's prediction year. They do not update with target-year outcomes.

\subsection{Pretraining, retrieval, and adaptation}

Terrain consists of elevation mean, standard deviation, minimum, maximum, and mean slope, aggregated from the approximately 4-km GridMET terrain grid to GFED cells. Cell centers receive EPA Level III categories by polygon membership, with nearest-polygon fallback when necessary. The category index is embedded in four dimensions inside the TabNet predictor. All other inputs are standardized using the appropriate historical pretraining period.

Let $J$ contain continuous feature indices other than history length. The model-independent retrieval representation is
\begin{equation}
v(u)=\left[(u_J-\mu_J)/s_J^{\mathrm{scale}},\;\operatorname{onehot}(e)\right],
\qquad d(u,u')=\|v(u)-v(u')\|_2.
\label{eq:ca-retrieval-distance}
\end{equation}
The division is elementwise. Here $\mu_J$ and $s_J^{\mathrm{scale}}$ are the fitted training scaler's mean and positive scale values: a population-standard-deviation estimate (ddof zero), with scale one for constant or numerically constant fields. With 25 continuous fields and a four-dimensional categorical embedding, the predictor expands 26 raw fields to 29 internal inputs; retrieval omits history length and uses 24 continuous fields plus 13 one-hot categories (37 dimensions). Its one-hot contribution to squared distance is $2\mathbf{1}\{e\ne e'\}$. There is no learned embedding distance, regional restriction, distance threshold, or retrieval-weight tuning. The primary union of exact $k=5$ neighbors over all 243,090 unlabeled target rows gives 363,643 unique historical rows with prevalence 0.075907. Within each of model seeds 11, 22, and 33, all ranking policies and continued focal use the same retrieved historical examples and frozen pretrained model. Sampled positive--negative pair identities agree across the three ranking methods over common epochs; different stopping epochs still imply different total update counts.

Pretrained models use the same TabNet widths/steps as Yosemite, Adam at $0.005$, zero weight decay, and a 120-epoch spatial cap. The focal loss for a set with prevalence $\pi$ is
\begin{equation}
\ell_{\mathrm{focal}}(s,y)=-w_y(1-p_y)^2\log p_y,
\qquad p_y=\sigma((2y-1)s),\quad w_1=1-\pi,\;w_0=\pi.
\label{eq:ca-focal}
\end{equation}
The continued-focal control recomputes these weights from the retrieved set. For direct, RDPO, and selective ranking, focal loss is detached and logged only; its coefficient is zero and it supplies no gradient. All local policies use AdamW at $10^{-4}$, weight decay 0.01, and batches of 4096. The epoch cap is twice the actual pretraining epoch count, with a minimum 15 epochs. The stopping rule compares adjacent five-epoch means of training loss; the absolute relative change must be below 0.001 for two consecutive checks. It monitors pretraining focal loss or the local total objective. For ranking adaptation this is the pairwise objective plus the zero-valued mask term; the monitored focal loss does not select stopping epochs.

The final spatial pretrained models for seeds 11 and 33 reach the 120-epoch cap without a recorded plateau pass; seed 22 passes the rule. Direct ranking stops after 77/158/76 epochs for seeds 11/22/33, RDPO after 126/158/64, and selective after 126/94/64. All nine ranking adaptations pass the local training-loss stopping rule. The single all-ones group-attention matrix makes the mask-loss term zero, so these experiments use the full input group rather than learning sparse spatial feature selection. Policy batch normalization is in training mode and pretrained-model batch normalization in evaluation mode, so matching initial weights need not produce identical scores.

The XGBoost baseline uses the same fixed sampled training rows. Its settings are 1,000 rounds, depth 6, learning rate 0.03, row and column sampling 0.8, minimum child weight 5, and L2 penalty 1. Its F1 threshold is selected on all sampled 2002--2021 training rows scored by the final model.

\subsection{Retrieval and source-selection ablation}
\label{app:retrieval-ablation}

We first isolate retrieval under continued focal adaptation across model seeds 11, 22, and 33 on the complete 666-cell daily grid. Each policy starts from its seed-matched pretrained checkpoint and receives exactly 6,800 AdamW updates with batch size 4,096, learning rate $10^{-4}$, and weight decay 0.01. The spatial-KNN and uniformly sampled random sets both contain 363,643 historical rows; full-archive adaptation uses all 972,892 rows. Fixing both selection size for the matched comparison and optimizer updates avoids confounding target-conditioned selection with data volume or compute. The frozen seed-matched sources receive no adaptation.

\begin{table}[t]
\centering
\captionsetup{skip=5pt}
\caption{California retrieval ablation under continued focal adaptation on 2022. Values are mean $\pm$ sample SD over seeds 11, 22, and 33. Each adapted policy starts from its seed-matched pretrained checkpoint and receives exactly 6,800 optimizer updates; the frozen source receives none. Random is sampled uniformly without replacement and exactly matches the spatial-KNN set size. Metrics use the complete daily grid and are percentages. Bold marks the best displayed value.}
\label{tab:california-focal-retrieval-multiseed}
\begingroup
\papertablefont
\setlength{\tabcolsep}{3.0pt}
\renewcommand{\arraystretch}{1.02}
\begin{tabularx}{\linewidth}{@{}>{\raggedright\arraybackslash}Xlrrrr@{}}
\toprule
\textbf{Historical data} & \textbf{Rows} & \textbf{Daily AP} & \textbf{R@5\%} & \textbf{Global AP} & \textbf{AUROC} \\
\midrule
Frozen pretrained source & -- & 21.10$\pm$1.24 & 17.98$\pm$1.36 & 19.30$\pm$0.63 & 78.34$\pm$0.37 \\
Full historical archive & 972,892 & 21.19$\pm$0.40 & 18.21$\pm$0.96 & 19.43$\pm$0.41 & 78.42$\pm$0.33 \\
Random matched subset & 363,643 & 20.88$\pm$0.53 & 17.90$\pm$1.19 & 19.21$\pm$0.48 & 78.33$\pm$0.33 \\
Spatial one-hot KNN, $k=5$ & 363,643 & \textbf{21.55$\pm$0.42} & \textbf{18.44$\pm$0.88} & \textbf{19.99$\pm$0.36} & \textbf{78.57$\pm$0.31} \\
\bottomrule
\end{tabularx}
\endgroup
\end{table}

Spatial KNN improves daily AP over the matched random subset in every seed. The paired mean gain is 0.66 percentage points (95\% $t$-interval $[0.29,1.04]$); relative to the full archive it is 0.36 points ($[0.01,0.71]$). Mean Top5\% recall, global AP, and AUROC also favor spatial KNN over random, but their three-seed paired intervals include zero. Spatial adaptation does not uniformly improve on the frozen source: its mean daily-AP gain is 0.45 points, but seed 11 decreases by 0.53 points and the paired interval includes zero. Thus target-conditioned spatial selection has reproducible daily-AP value relative to matched random and full-history focal adaptation under this fixed budget, while the evidence does not establish a uniform improvement over no adaptation or across all metrics.

\subsection{Thresholds and independent metric checks}

Final policy thresholds maximize F1 on all sampled 2002--2021 training rows scored by that policy. They are applied unchanged to the complete 2022 grid. No test outcomes select a threshold, checkpoint, or calibration map; no probability calibration is fitted. The saved sigmoid probabilities determine operating-point counts, while actual checkpoint logits determine all ranking metrics to avoid saturation ties.

The evaluation verifies common labels, dates, and coordinates, saved-probability agreement with recomputed logits, finite scores, and saved confusion counts. The reference checkpoint and retrieved-index hashes are checked for every ranking policy. Realized sampled preference-pair hashes match across the three objectives for the first 77, 94, and 64 epochs in seeds 11, 22, and 33, respectively. Different stopping epochs imply different total update counts. Tables report mean and sample SD over these three model seeds on the same test year; this variability concerns optimization randomness.

\subsection{Pooled and daily evaluation metrics}

Global AP and AUROC pool all 243,090 grid--days. Pooled R@$q$ selects $\lceil qN\rceil$ highest raw scores across all dates and divides captured positives by 14,760, for $q=0.20$ and $0.30$, using the same partial-sort convention as Yosemite. Daily Top5\%, Top20\%, and Top30\% instead select 34, 134, and 200 cells separately per date, with score ties ordered by latitude then longitude. Daily AP, recall, and NDCG average the 351 positive dates. Total FN counts uncaptured positive cell--days and is distinct from one minus macro daily recall. Threshold F1 is an auxiliary operating-point diagnostic.

For the raw-DM severity analysis in Figure~\ref{fig:california-raw-dm-recall-gain}, we restrict evaluation to the 2,193 cell--days with observed $D_{t,j}>0$, order them by $D_{t,j}$, and take exactly $\lceil q\,2193\rceil$ examples for $q\in\{0.05,0.10,0.20\}$. This gives 110, 220, and 439 cell--days, with boundary values $9.2034816\times10^9$, $3.885151744\times10^9$, and $2.177735936\times10^9$, respectively; none of these boundaries has a tie. Recall is the fraction whose model probability exceeds that policy's training-selected threshold. The plotted quantities are seed-wise recall differences from continued focal, summarized by their mean and sample SD. This diagnostic is retrospective: 2022 dry matter defines the severity strata but never enters model fitting, threshold selection, or deployment-time ranking.

\begin{table}[t]
\centering
\caption{Complete-grid Top5\% selection: 34 cells per date and 12,410 alarms per policy. Counts are neighborhood-smoothed positive cell--days, not independent incidents. Means over three seeds. Ranking adaptation has no focal term; paired gains use same-seed continued focal.}
\label{tab:california-fulltarget-budget}
\begingroup
\papertablefont
\setlength{\tabcolsep}{2pt}
\renewcommand{\arraystretch}{1.08}
\begin{tabularx}{\linewidth}{@{}l*{3}{>{\centering\arraybackslash}X}@{}}
\toprule
Method & FN & Captured & Paired gain \\
\midrule
Continued focal & \mbox{11894.00} & \mbox{2866.00} & \mbox{0.00} \\
Direct ranking & \mbox{11583.33} & \mbox{3176.67} & \mbox{310.67} \\
RDPO & \mbox{11593.33} & \mbox{3166.67} & \mbox{300.67} \\
Selective ranking & \mbox{\textbf{11549.67}} & \mbox{\textbf{3210.33}} & \mbox{\textbf{344.33}} \\
\bottomrule
\end{tabularx}
\endgroup
\end{table}

\begin{table}[t]
\centering
\caption{California 2022 complete-grid evaluation, 666 cells per date. Ranking adaptation excludes focal supervision. Mean $\pm$ sample SD over seeds 11, 22, 33; score metrics are percentages. R@20/R@30 pool raw scores across dates, matching Yosemite. Recall/F1 use each policy threshold selected on training data.}
\label{tab:california-pooled-ranking}
\begingroup
\fontsize{8}{10}\selectfont
\setlength{\tabcolsep}{2pt}
\renewcommand{\arraystretch}{1.08}
\begin{tabularx}{\linewidth}{@{}>{\raggedright\arraybackslash}Xrrrrrr@{}}
\toprule
Method & AUPRC & AUROC & R@20 & R@30 & Recall & F1 \\
\midrule
XGBoost & \mbox{11.02\,$\pm$\,0.02} & \mbox{66.27\,$\pm$\,0.04} & \mbox{38.38\,$\pm$\,0.03} & \mbox{51.45\,$\pm$\,0.06} & \mbox{31.54\,$\pm$\,0.86} & \mbox{17.81\,$\pm$\,0.07} \\
Pretrained spatial focal & \mbox{19.30\,$\pm$\,0.63} & \mbox{78.34\,$\pm$\,0.37} & \mbox{54.60\,$\pm$\,0.61} & \mbox{68.55\,$\pm$\,0.41} & \mbox{46.51\,$\pm$\,1.58} & \mbox{26.34\,$\pm$\,0.69} \\
Continued focal & \mbox{20.07\,$\pm$\,0.34} & \mbox{78.60\,$\pm$\,0.21} & \mbox{55.09\,$\pm$\,0.64} & \mbox{68.70\,$\pm$\,0.93} & \mbox{45.70\,$\pm$\,1.77} & \mbox{\textbf{26.58}\,$\pm$\,0.14} \\
Direct ranking & \mbox{20.80\,$\pm$\,1.53} & \mbox{79.46\,$\pm$\,0.50} & \mbox{\textbf{56.41}\,$\pm$\,0.92} & \mbox{\textbf{70.51}\,$\pm$\,0.91} & \mbox{\textbf{54.80}\,$\pm$\,5.07} & \mbox{26.52\,$\pm$\,0.31} \\
RDPO & \mbox{\textbf{20.89}\,$\pm$\,1.39} & \mbox{\textbf{79.47}\,$\pm$\,0.47} & \mbox{56.24\,$\pm$\,0.91} & \mbox{70.42\,$\pm$\,0.97} & \mbox{54.71\,$\pm$\,5.26} & \mbox{26.42\,$\pm$\,0.41} \\
Selective ranking & \mbox{20.71\,$\pm$\,1.13} & \mbox{79.38\,$\pm$\,0.33} & \mbox{56.16\,$\pm$\,0.78} & \mbox{70.10\,$\pm$\,0.42} & \mbox{54.20\,$\pm$\,4.62} & \mbox{26.43\,$\pm$\,0.43} \\
\bottomrule
\end{tabularx}
\endgroup
\end{table}

\begin{table}[t]
\centering
\caption{California 2022 complete-grid evaluation, 666 cells per date. Ranking adaptation excludes focal supervision. Mean $\pm$ sample SD over seeds 11, 22, 33; score metrics are percentages. Metrics average the 351 positive dates; daily selection uses 34 cells.}
\label{tab:california-daily-ranking}
\begingroup
\papertablefont
\setlength{\tabcolsep}{2pt}
\renewcommand{\arraystretch}{1.08}
\begin{tabularx}{\linewidth}{@{}l*{3}{>{\centering\arraybackslash}X}@{}}
\toprule
Method & Daily AP & R@Top5\% & NDCG@Top5\% \\
\midrule
XGBoost & \mbox{14.24\,$\pm$\,0.07} & \mbox{11.64\,$\pm$\,0.23} & \mbox{16.60\,$\pm$\,0.26} \\
Pretrained spatial focal & \mbox{21.10\,$\pm$\,1.24} & \mbox{17.98\,$\pm$\,1.36} & \mbox{26.16\,$\pm$\,2.08} \\
Continued focal & \mbox{21.62\,$\pm$\,0.33} & \mbox{18.70\,$\pm$\,0.74} & \mbox{27.06\,$\pm$\,0.81} \\
Direct ranking & \mbox{24.35\,$\pm$\,0.69} & \mbox{22.25\,$\pm$\,1.39} & \mbox{31.05\,$\pm$\,1.44} \\
RDPO & \mbox{24.47\,$\pm$\,0.42} & \mbox{22.11\,$\pm$\,1.92} & \mbox{31.15\,$\pm$\,1.47} \\
Selective ranking & \mbox{\textbf{24.57}\,$\pm$\,0.52} & \mbox{\textbf{22.79}\,$\pm$\,2.07} & \mbox{\textbf{31.63}\,$\pm$\,1.75} \\
\bottomrule
\end{tabularx}
\endgroup
\end{table}

\begin{table}[t]
\centering
\caption{California 2022 complete-grid evaluation, 666 cells per date. Ranking adaptation excludes focal supervision. Mean $\pm$ sample SD over seeds 11, 22, 33; score metrics are percentages. Daily budgets select 34, 134, and 200 cells, respectively; recall averages positive dates.}
\label{tab:california-daily-budgets}
\begingroup
\papertablefont
\setlength{\tabcolsep}{2pt}
\renewcommand{\arraystretch}{1.08}
\begin{tabularx}{\linewidth}{@{}l*{3}{>{\centering\arraybackslash}X}@{}}
\toprule
Method & R@Top5\% & R@Top20\% & R@Top30\% \\
\midrule
Direct ranking & \mbox{22.25\,$\pm$\,1.39} & \mbox{\textbf{56.82}\,$\pm$\,1.47} & \mbox{\textbf{71.49}\,$\pm$\,0.66} \\
RDPO & \mbox{22.11\,$\pm$\,1.92} & \mbox{56.41\,$\pm$\,1.34} & \mbox{71.27\,$\pm$\,0.74} \\
Selective ranking & \mbox{\textbf{22.79}\,$\pm$\,2.07} & \mbox{56.21\,$\pm$\,1.36} & \mbox{71.22\,$\pm$\,0.65} \\
\bottomrule
\end{tabularx}
\endgroup
\end{table}

Selective's Top5\% advantage over RDPO is concentrated in seed 22, where the recall difference is 1.97 points; the other two differences are approximately zero and 0.07 points. This supports a small-budget prioritization result without a uniform advantage across seeds or budgets. The seed-22 selective and RDPO policies stop at 94 and 158 epochs, respectively; this comparison includes their different realized training durations.

\subsection{Held-out pair strata and equal-budget misses}

The pair audit enumerates all positive--negative pairs within each complete-grid 2022 date. The frozen seed-specific pretrained gap defines correct ($\Delta_{\rm ref}>0$) and wrong ($\Delta_{\rm ref}<0$) strata; ties are excluded from both. Correction and retention require $\Delta_\theta>0$, and expansion requires $\Delta_\theta>\Delta_{\rm ref}$ on reference-correct pairs. These are evaluation diagnostics, not additional training supervision or guarantees on unseen pairs.

\begin{table}[t]
\centering
\caption{California 2022 complete-grid evaluation, 666 cells per date. Ranking adaptation excludes focal supervision. Mean $\pm$ sample SD over seeds 11, 22, 33; score metrics are percentages. Each seed enumerates 8,865,724 same-date positive--negative pairs. Correction/retention require a positive policy gap; expansion requires a larger gap than the frozen reference on reference-correct pairs. Reference ties are excluded from both strata.}
\label{tab:california-pair-strata}
\begingroup
\papertablefont
\setlength{\tabcolsep}{2pt}
\renewcommand{\arraystretch}{1.08}
\begin{tabularx}{\linewidth}{@{}l*{3}{>{\centering\arraybackslash}X}@{}}
\toprule
Method & Wrong corrected & Correct retained & Correct expanded \\
\midrule
Direct ranking & \mbox{33.14\,$\pm$\,2.46} & \mbox{93.08\,$\pm$\,0.99} & \mbox{92.62\,$\pm$\,1.09} \\
RDPO & \mbox{32.93\,$\pm$\,3.40} & \mbox{\textbf{93.17}\,$\pm$\,0.98} & \mbox{\textbf{92.72}\,$\pm$\,1.10} \\
Selective ranking & \mbox{\textbf{33.66}\,$\pm$\,3.32} & \mbox{92.90\,$\pm$\,0.64} & \mbox{92.43\,$\pm$\,0.72} \\
\bottomrule
\end{tabularx}
\endgroup
\end{table}

\begin{table}[t]
\centering
\caption{California 2022 complete-grid evaluation, 666 cells per date. Ranking adaptation excludes focal supervision. Mean $\pm$ sample SD over seeds 11, 22, 33; score metrics are percentages. Daily Top5\% uses 12,410 alarms against 14,760 positives. Zero-hit dates have positives but no captured positive.}
\label{tab:california-equal-budget-misses}
\begingroup
\papertablefont
\setlength{\tabcolsep}{2pt}
\renewcommand{\arraystretch}{1.08}
\begin{tabularx}{\linewidth}{@{}l*{3}{>{\centering\arraybackslash}X}@{}}
\toprule
Method & FN & FP & Zero-hit dates \\
\midrule
XGBoost & \mbox{13028.00\,$\pm$\,20.78} & \mbox{10678.00\,$\pm$\,20.78} & \mbox{87.33\,$\pm$\,1.53} \\
Pretrained spatial focal & \mbox{11985.00\,$\pm$\,142.43} & \mbox{9635.00\,$\pm$\,142.43} & \mbox{70.67\,$\pm$\,8.50} \\
Continued focal & \mbox{11894.00\,$\pm$\,60.26} & \mbox{9544.00\,$\pm$\,60.26} & \mbox{67.00\,$\pm$\,6.93} \\
Direct ranking & \mbox{11583.33\,$\pm$\,101.48} & \mbox{9233.33\,$\pm$\,101.48} & \mbox{40.33\,$\pm$\,6.03} \\
RDPO & \mbox{11593.33\,$\pm$\,131.46} & \mbox{9243.33\,$\pm$\,131.46} & \mbox{42.67\,$\pm$\,13.87} \\
Selective ranking & \mbox{\textbf{11549.67}\,$\pm$\,159.86} & \mbox{\textbf{9199.67}\,$\pm$\,159.86} & \mbox{\textbf{39.67}\,$\pm$\,15.89} \\
\bottomrule
\end{tabularx}
\endgroup
\end{table}

\section{Design for larger-area deployment: untested extensions}
\label{app:broader-design}

The California pilot combines spatial risk pretraining, target-conditioned retrieval, and ranking adaptation without focal supervision. The following untested design describes ways to adapt this pretrained foundation for larger-area deployment; the reported experiments use the architecture and protocol in Appendix~\ref{app:california}.

\paragraph{Partial pooling of spatial effects.}
A shared weather/landscape backbone can be complemented by a region residual,
\begin{equation}
r_{t,i}=f_0(w_{t-1,i},a_i,h_{\operatorname{yr}(t),i},d_t)
+b_{e_i}+v_{e_i}^{\top}\psi(w_{t-1,i},a_i),
\label{eq:broader-hierarchical-risk}
\end{equation}
where $b_e$ represents regional baseline differences and $v_e$ regional response differences. A penalty $\rho\sum_e(b_e^2+\|v_e\|_2^2)$ with $\rho>0$ partially pools poorly supported regions toward the shared model. This permits regularized regional intercept and response corrections without fitting independent predictors for every county. Their decomposition is generally nonunique when the shared model and regional effects are freely learned together. For an interpretable staged fit, hold $f_0$ and $\psi$ fixed, center/normalize $\psi$ on historical training rows, and impose $\sum_e n_e b_e=0$ and $\sum_e n_e v_e=0$ with fixed support weights $n_e>0$. These conventions do not establish a causal interpretation. Neither this explicit head nor its penalty is implemented in the current TabNet embedding model. Unknown-region effects default to the shared component; they cannot be learned from a region's future labels.

\paragraph{Retrieval support and adaptive distance.}
A candidate distance family is
\begin{align}
d_{\omega}^2(u,u')={}&
\omega_w\|\widetilde w-\widetilde w'\|^2+
\omega_a\|\widetilde a_{\mathrm{land}}-\widetilde a'_{\mathrm{land}}\|^2+
\omega_h\|\widetilde h-\widetilde h'\|^2\notag\\
&+\omega_d\|d_t-d_{t'}\|^2+
\omega_e\mathbf{1}\{e\ne e'\}+\omega_{\mathrm{geo}}[d_{\mathrm{geo}}(i,i')/L]^2,
\quad \omega_\cdot\ge0.
\label{eq:broader-distance}
\end{align}
Here $i,i'$ are the cells associated with $u,u'$. The displayed squared quantity is a weighted dissimilarity. Its square root $d_\omega$ is a pseudometric on the represented inputs, and is a metric only if the positively weighted components distinguish them; zero weights can remove those distinctions. Here $a_{\mathrm{land}}$ excludes coordinates, $d_{\mathrm{geo}}$ is a geodesic or suitable projected distance, and $L>0$ sets its scale. Select weights, support thresholds, and $k$ on historical folds before the final test; compare against the pilot's unweighted standardized distance. Combining ecological similarity with weather similarity permits borrowing across regions while reducing unsupported neighbors. A fallback to the frozen global model when historical support is insufficient must be specified before evaluating its effect. The pilot's one-hot category term corresponds to a fixed categorical distance, not a learned metric or a support-aware fallback.

\paragraph{Within-region and cross-region preferences.}
A regional sampling study can use
\begin{equation}
\mathcal{L}_{\mathrm{spatial\ pair}}
=(1-\xi)\,\mathbb{E}_{q_{\mathrm{within}}}\ell(\Delta_\theta-\tau)
+\xi\,\mathbb{E}_{q_{\mathrm{cross}}}\ell(\Delta_\theta-\tau),
\quad \ell(m)=\operatorname{softplus}(-\beta_{\mathrm{rank}}m)/\beta_{\mathrm{rank}},
\label{eq:broader-pair-mixture}
\end{equation}
with $0\le\xi\le1$, $\beta_{\mathrm{rank}}>0$, and normalized same-date positive--negative pair distributions in both terms. Within-region pairs test local weather response; cross-region pairs connect score scales needed for area-wide prioritization. Regions or dates without both labels require an explicit skip-or-borrow rule. The pilot allows both pair types naturally but has no tuned mixture or regional reweighting. A fair loss comparison holds the sampled pairs, initialization, batches, and stopping rule fixed and selects each loss weight on historical folds; an equal fixed weight is only a diagnostic comparison.

\paragraph{Calibration with sparse-region shrinkage.}
If historical calibration diagnostics show persistent regional errors, a possible hierarchical map is
\begin{equation}
\widehat p_{t,i}=\sigma(a_{e_i}s_\theta(u_{t,i})+b_{e_i}),
\qquad a_e=\operatorname{softplus}(\alpha_0+\delta_e),\quad b_e=b_0+\gamma_e,
\label{eq:broader-regional-platt}
\end{equation}
fitted to strictly forward historical scores with shrinkage on $\delta_e,\gamma_e$. Obtain such scores through separate adaptation at each issue year using only information available then; pooling later-year histories can expose earlier calibration outcomes. Center the deviations with $\sum_e n_e\delta_e=\sum_e n_e\gamma_e=0$ to fix the common-effect convention. Positive slopes preserve order \emph{within} a region, but region-specific maps can change \emph{cross-region} order. Therefore they require a new area-wide ranking evaluation, because the ordering invariance of a single global monotone map does not extend to region-specific maps. Use one global monotone map when preserving the original statewide ranking is part of the deployment requirement. Keep the ranking loss in its score space unless a separately tested objective explicitly optimizes calibrated probabilities.

\paragraph{Staged, deployment-matched evaluation.}
First hold the shared backbone fixed and ablate coordinates, terrain, causal rates, and ecological embeddings, giving baselines the same historical information access. Then compare continued focal, direct, RDPO, and selective ranking on several chronological windows with historical-only selection of coefficients. Report pooled/daily AP, recall and FN at the same selected fraction, regional support, within/cross-region pair accuracy, and calibration both overall and by region. Separately test unseen locations through spatial blocks or held-out ecological regions \citep{roberts2017crossvalidation}. Historical rates from a held-out region are legitimate only for a deployment scenario in which its past labels are available; a zero-shot geographic test must remove that information and provide an unknown-category fallback. Temporal and spatial uncertainty need date/spatial blocks, not independent row resampling.

Finally, replace annual-batch retrieval with rolling covariate windows for an online forecasting claim. At an issue time, all retrieval inputs, label histories, embedding vocabulary, and calibration fits must be available then. Evaluate the full daily candidate grid at the intended physical smoothing scale and selected fraction.

\section{Discussion and limitations}
\label{sec:discussion}
\label{app:discussion}

\paragraph{Pretraining as a foundation for adaptation.}
The evaluated framework uses a historically trained risk model as the starting point for subsequent fine-tuning. All local ranking variants inherit its parameters and representation. Direct ranking optimizes positive--negative score gaps; RDPO and selective use frozen pretrained gaps to define alternative loss targets. California compares these targets at fixed pairwise settings, while Yosemite compares binary-label and cross-input preferences. Both experiments exclude focal supervision from ranking adaptation; their pair contexts, temperatures, and stopping schedules remain dataset-specific. The experiments support ranking adaptation from the evaluated checkpoints. The fixed-update continued-focal ablation isolates source selection and supports a daily-AP benefit over matched random and full-history adaptation. It does not establish that retrieval is necessary, nor isolate its interaction with the ranking loss. Staging and pretrained-model quality are not separately controlled. The scalar-gradient analysis characterizes objective behavior under its stated assumptions, rather than establishing a generalization guarantee.

\paragraph{Deployment covariates and information access.}
Retrieval uses unlabeled deployment covariates, so the evaluation is batch-transductive. California's primary protocol uses covariates from the entire target year before adaptation and evaluates all 666 cells on every test date. Although each row uses previous-day weather, adaptation sees the future target covariate distribution. Daily budgets specify the evaluation decision, not an online information protocol; issue-time adaptation using only then-available covariates remains unevaluated. The spatial pretrained model receives 26 raw fields, including an ecological category represented by a learned embedding, and its causal history features draw on full historical label grids from prior years.

\paragraph{Evaluation scope.}
California evaluates one target year on locations present historically, so spatial heterogeneity here concerns temporal adjustment on a fixed grid; unseen-region generalization and spatial feature contributions are not isolated. The reported seed variation measures training randomness conditional on the evaluated years. The three ranking objectives use a common training-loss stopping rule but different realized update counts, which limits attribution of small differences solely to their targets. Fixed $k$ and dataset-specific ranking temperatures are evaluated configurations, rather than demonstrated optima from a historical-forward sensitivity study. Yosemite's fixed joint-training comparison changes data access, optimizer, and update counts, so it does not isolate initialization alone.

\paragraph{Probability interpretation and calibration.}
Modern neural networks are often poorly calibrated \citep{guo2017calibration}. The no-focal ranking objectives optimize score gaps without constraining probabilities to be calibrated, so calibration is distinct from the ranking evaluation reported here. All reported California policies are evaluated without a fitted calibration map. Deployment requiring probability estimates should fit and evaluate calibration on strictly forward historical predictions, separately from ranking assessment.

\paragraph{High-activity diagnostic.}
High raw-DM strata are defined retrospectively for evaluation; training uses binary neighborhood-smoothed occurrence labels. The reported recall gains use policy-specific historical F1 thresholds and do not control alarm rates or demonstrate severity calibration. They support sensitivity to high observed activity under that threshold protocol. Positive cell--days may share the same underlying fire and should not be interpreted as counts of independent incidents.

\end{document}